\documentclass[11pt]{article}

\usepackage[english]{babel}

\usepackage[a4paper, top=2cm, bottom=2cm, left=2cm, right=2cm, marginparwidth=1.75cm]{geometry}

\usepackage{graphicx}
\usepackage{caption}
\usepackage{subcaption}
\usepackage[colorlinks=true, allcolors=blue]{hyperref}
\usepackage[utf8]{inputenc}
\usepackage{amsmath,amssymb}
\usepackage{multirow}
\usepackage{parskip}
\usepackage[nottoc]{tocbibind}
\usepackage{enumitem}
\usepackage{pifont}

\numberwithin{equation}{section}
\DeclareMathOperator{\Tr}{Tr}

\title{Graph Neural Networks for Airfoil Design}
\author{Florent Bonnet}
\date{August 31, 2021}

\begin{document}
\maketitle

\tableofcontents

\newpage

\begin{abstract}
The study of partial differential equations (PDE) through the framework of deep learning has emerged a few years ago leading to the impressive approximations of simple dynamics. Graph neural networks (GNN) turned out to be very useful in those tasks by allowing the treatment of unstructured data often encountered in the field of numerical resolutions of PDE. Howewver, the resolutions of harder PDE such as Navier-Stokes equations are still a challenging task and most of the works done on the latter concentrate either on simulating the flow around simple geometries or on qualitative results that looks physical for design purpose. In this study, we try to leverage the work done on deep learning for PDE and GNN by proposing an adaptation of a known architecture in order to tackle the task of approximating the solution of the two dimensional steady-state incompressible Navier-Stokes equations over different airfoil geometries. In addition to that, we test our model not only on its performance over the volume but also on its performance to approximate surface quantities such as the wall shear stress or the isostatic pressure leading to the inference of global coefficients such as the lift and the drag of our airfoil in order to allow design exploration. This work takes place in a longer project that aims to approximate three dimensional steady-state solutions over industrial geometries.
\end{abstract}

\section{Introduction} \label{introduction}

The study of dynamical systems through neural networks is a new paradigm that allow us to see the resolution of Ordinary Differential Equations (ODE) or Partial Differential Equations (PDE) under a new light. The classical study of numerical resolution of such equations use different tools such as the finite-element method or the finite-volume method to discretize the differential operator and to solve the equations through the inversion of a linear operator. Such techniques are powerful but need to balance the accuracy desired and the time or resources available for the task. With the recent rise of Deep Learning \cite{Goodfellow-et-al-2016} and the availability of more and more datas and simulations, new techniques have grown to tackle the resolution of differential equations. Various strategies have been proposed such as the direct approximation of the infinitesimal dynamic \cite{neuralODE}, fix point methods mimicking infinite-depth networks \cite{DEQ}, variational principles and conservative systems \cite{HNN}, the direct minimization of the differential equation residue \cite{pinns} or also more generic data-driven frameworks to approximate mappings between Hilbert spaces \cite{deeponet, GKN}. Those methods allow us, after a certain training time, to have access to new accurate solutions almost instantly which completely rebalances the accuracy-time compromise that holds in more classical techniques.

Moreover, classical simulation are often done on unstructured mesh that does not immediately fall in the historical task of Deep Learning. Graph neural networks (GNN) appeared recently \cite{GCN, SAGE, GAT} to tackle semi-supervised tasks such as protein interactions inference or interactions in social networks where a graph can be a priori built. Other networks such as PointNet \cite{pointnet} led to the classification of generic point clouds. Those tools are of great use for dealing with unstructured data and they necessarily have been used in the field of differential equations resolution leading to new ways of looking at it \cite{grand}.

In this work, we try to take advantage of those two frameworks to approximate the solution of the two dimensional steady-state incompressible Navier-Stokes equations on different airfoil geometries. This is a preliminary work to the harder task of three dimensional industrial simulations approximations. The goal is not only to have a good approximation of the different fields over the volume but also to approximate correctly surface quantities such as the wall shear stress or the isostatic pressure. In section \ref{datasets} we present the different datasets we use in this work and explain their differences, this is followed by a theoretical review in section \ref{hydro} and \ref{PDE} that presents the basis of fluid dynamics and some results in the theory of PDE. After this theoretical overview, we present, in section \ref{experiments}, our work and the architecture we propose. Finally, we end this report by a conclusion, some perspectives and acknowledgements in section \ref{conclusion} and \ref{merci}.

\subsubsection*{Toolbox}

All the coding part has been done in Python, the deep learning framework used is PyTorch \cite{pytorch} and the framework used for GNN is PyTorch Geometric \cite{geometric}. The tool we used for the simulations is OpenFoam \cite{foam} an open source software that contains multiple solvers. For the visualization and the manipulation of the data, we used the software ParaView \cite{paraview} and an associated Python library PyVista \cite{pyvista}. The computations are done on Google Cloud Platform with the help of an nVidia Tesla P100 16Go.

\subsubsection*{Extrality, MLIA and MVA}

The internship has been done at Extrality in collaboration with the MLIA laboratory of LIP6 in Sorbonne Université. My tutors were Pierre Yser and Ahmed Mazari at Extrality and Patrick Gallinari at MLIA. Extrality is a young start up that has its offices at Agoranov. It is born in 2019 and has now 12 employees. Its first goal is to develop a SaaS platform for businesses to allow them to run fluid dynamics simulations over new geometries almost instantly in order to let more freedom in industrial design. On the other hand, the MLIA is one of the Machine Learning laboratory of Sorbonne Université, their axes of research goes from classical task such as computer vision or natural language processing to dynamical systems. My tutor at the MVA master was Guillaume Charpiat of the TAO/TAU team at INRIA Saclay.

\section{Datasets presentation} \label{datasets}

In this work we will use six datasets, one for the training, one for the validation and four for the test. Each of the test sets have their particularity and are here to test the ability of the model to generalize in different directions from the training set distribution. Each of this datasets are a collection of different airfoil geometries from, for example, the National Advisory Committee for Aeronautics (NACA). We define a geometry as a shape of an airfoil, an angle of attack and an aspect ratio. For each of those geometries, we define a compact domain around it, build a mesh on this domain with the help of Gmsh and run an incompressible steady-state CFD simulation over it with the help of OpenFoam. More precisely, we use the solver SimpleFoam as it is dedicated to incompressible steady-state simulations. The \emph{Spallart-Allmaras} model is used in order to model the turbulence\footnotemark. Those simulations take in inputs the geometry, the mesh and the velocity at the inlet and output four quantities of interest that will be the targets in our supervised task, namely the $x$ and $y$ components of the velocity field, the pressure field and the turbulent viscosity field\footnotemark[\value{footnote}]. However, the incompressibility equation is not well verified in those simulations, it looks like a more recent version of OpenFoam solve this problem but we did not redo the simulations as it was not the core of the internship. Hence, we will not look at the incompressibility constraint in all of this work.

\begin{figure}
    \begin{subfigure}{.5\textwidth}
      \centering
      \includegraphics[width=\linewidth]{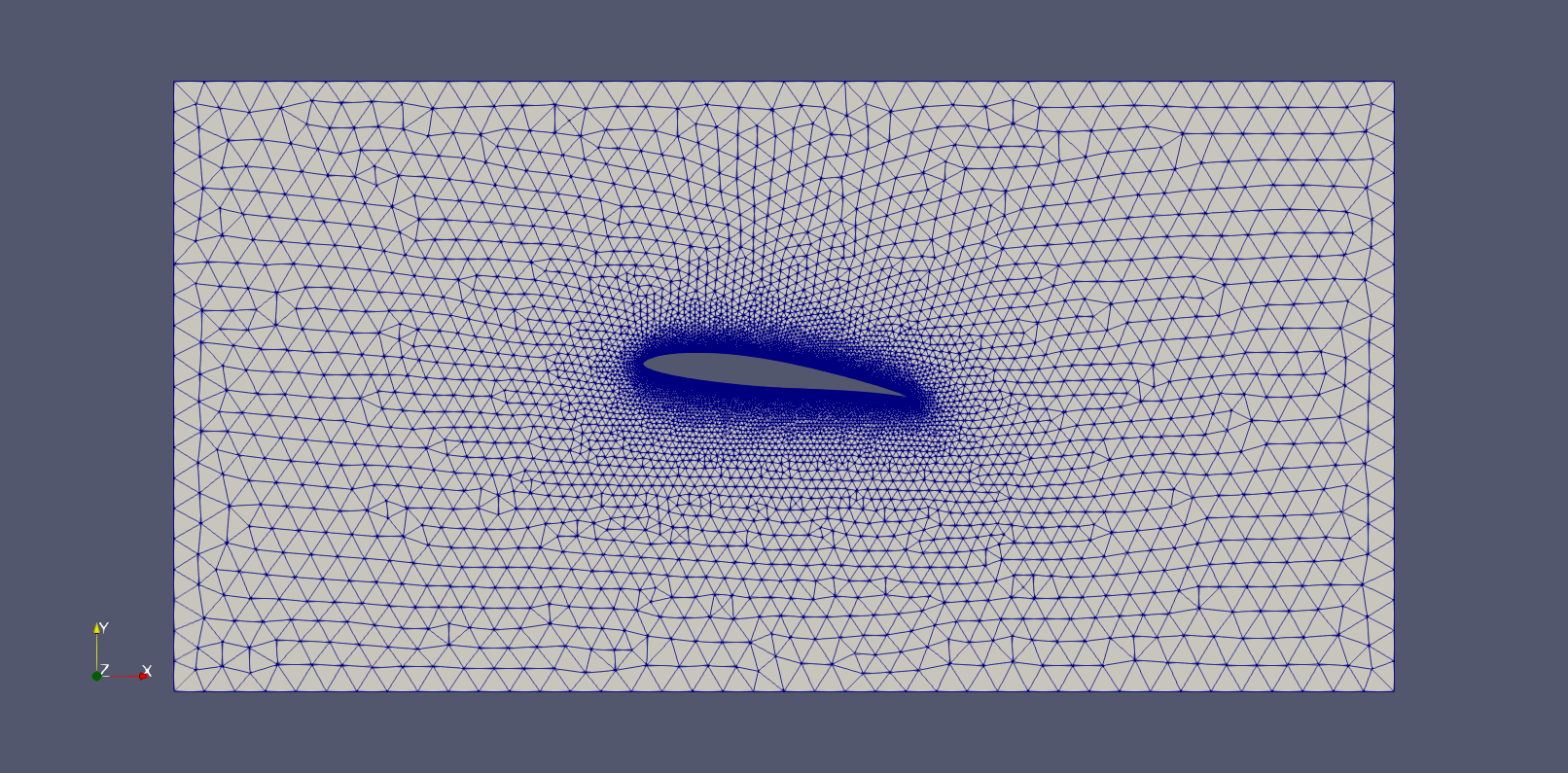}  
      \caption{Training set}
    \end{subfigure}
    \begin{subfigure}{.5\textwidth}
      \centering
      \includegraphics[width=\linewidth]{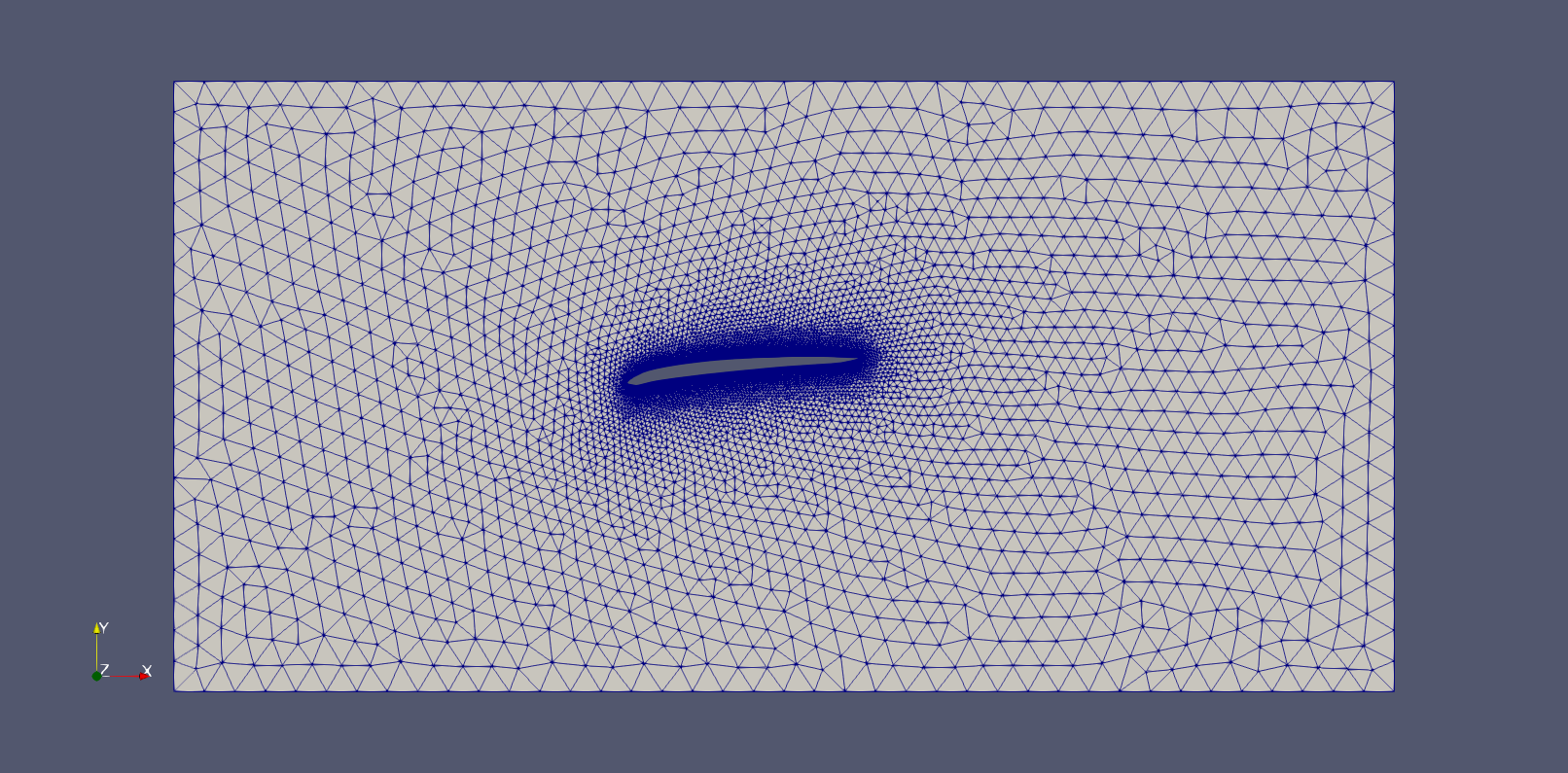}  
      \caption{Validation set}
    \end{subfigure}

    \begin{subfigure}{.5\textwidth}
      \centering
      \includegraphics[width=\linewidth]{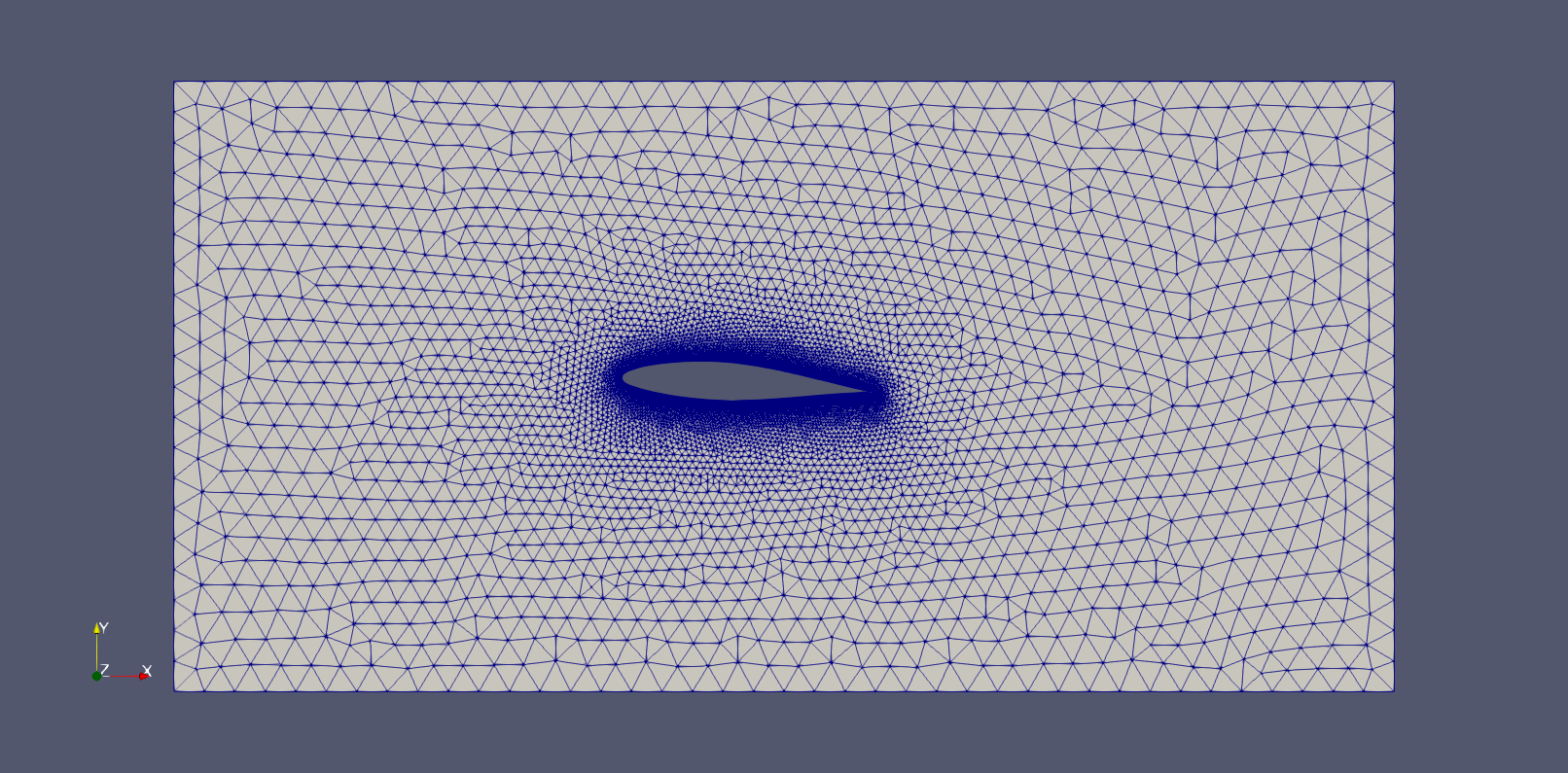}  
      \caption{Test set}
    \end{subfigure}
    \begin{subfigure}{.5\textwidth}
      \centering
      \includegraphics[width=\linewidth]{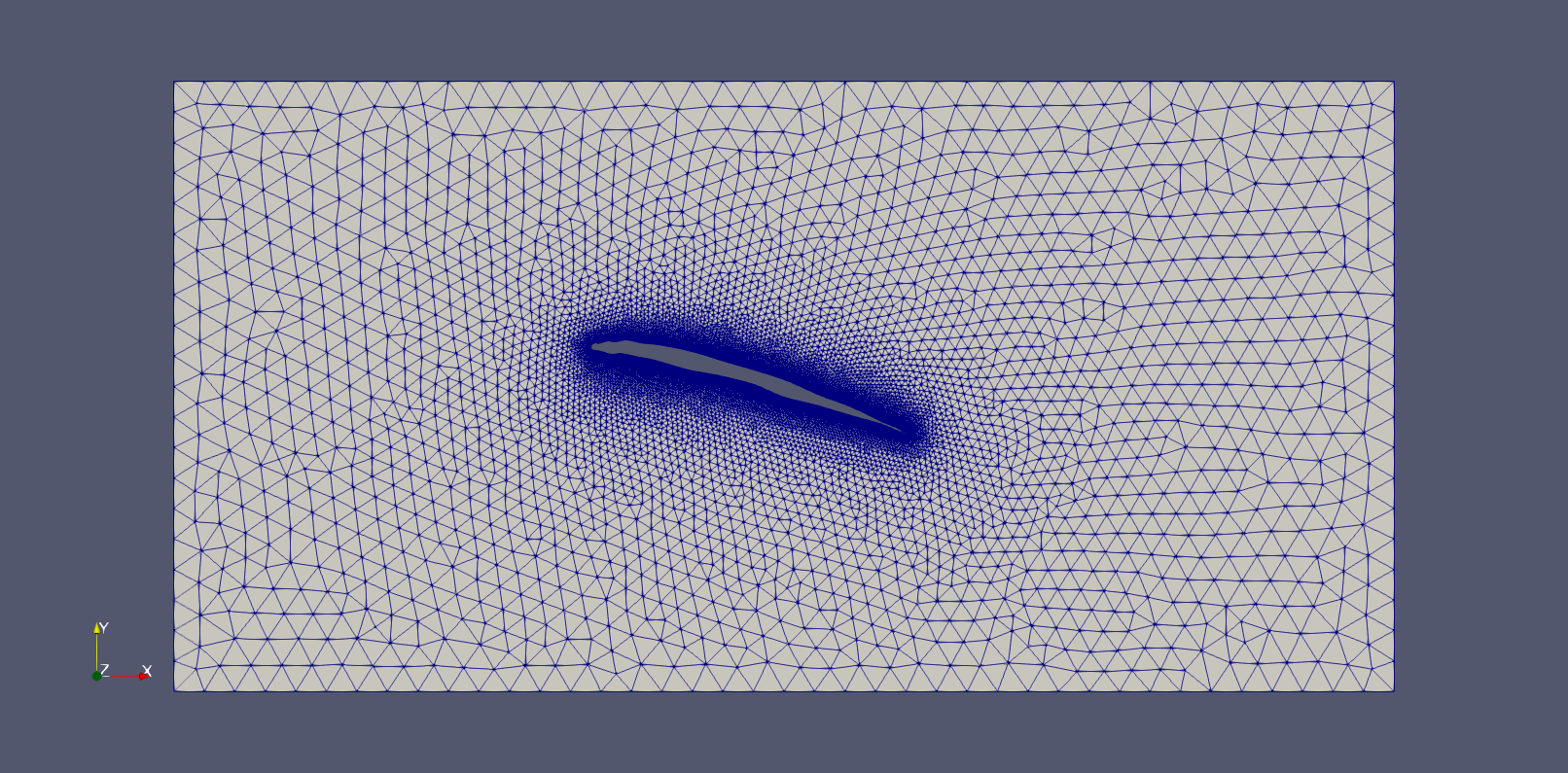}  
      \caption{Test noise set}
    \end{subfigure}
    
    \begin{subfigure}{.5\textwidth}
      \centering
      \includegraphics[width=\linewidth]{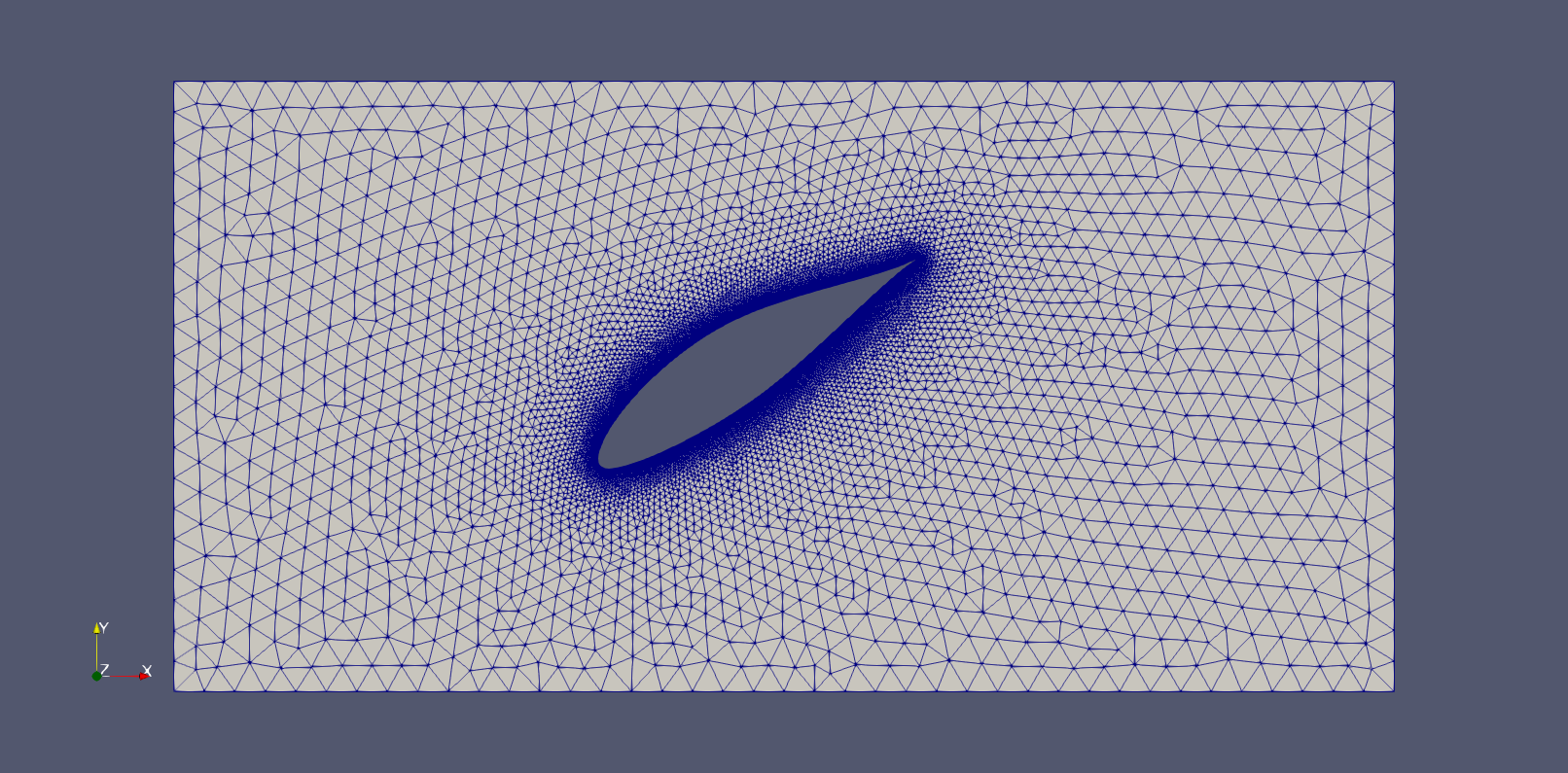}  
      \caption{Test rotation set}
    \end{subfigure}
    \begin{subfigure}{.5\textwidth}
      \centering
      \includegraphics[width=\linewidth]{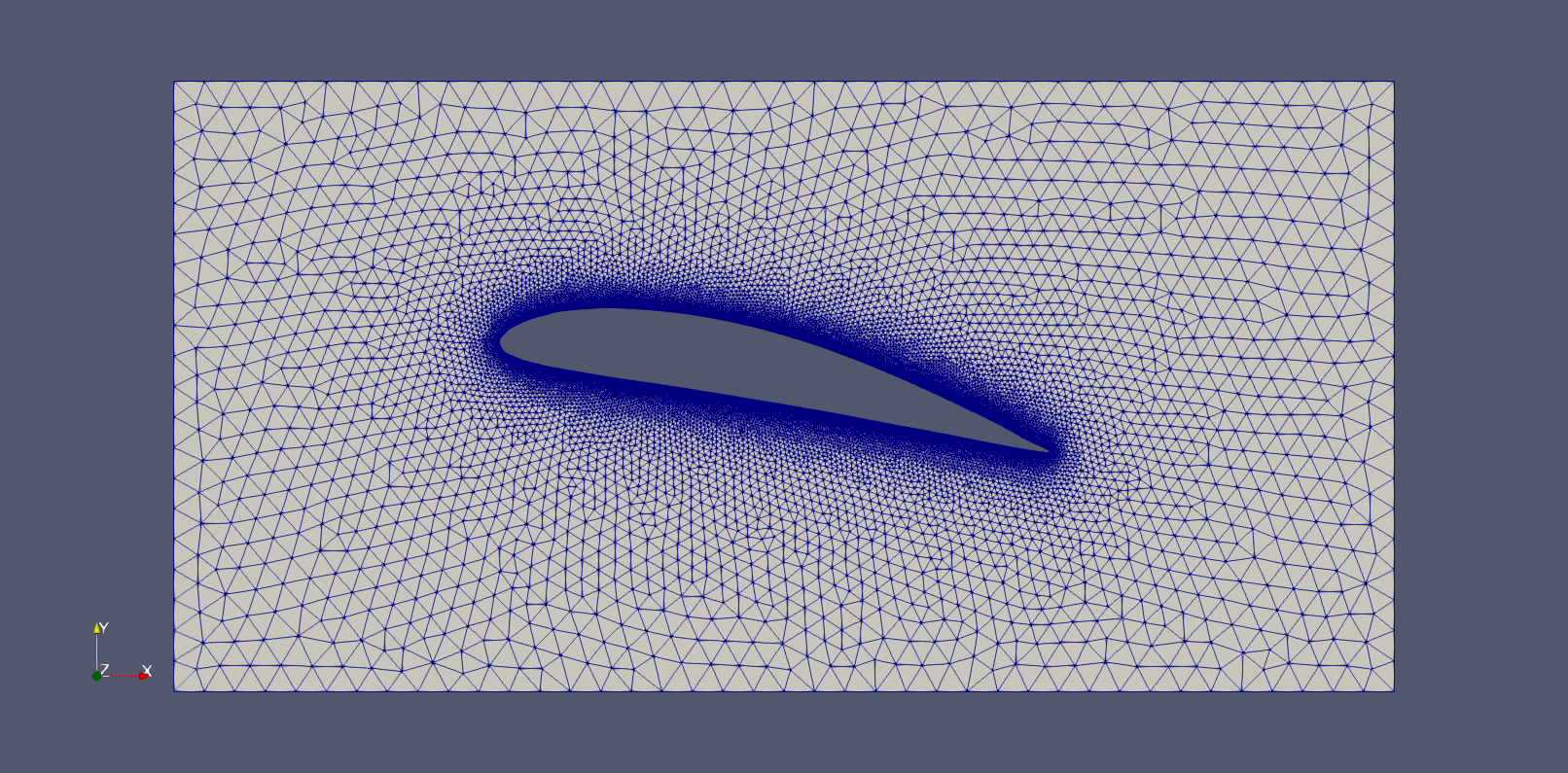}  
      \caption{Test big set}
    \end{subfigure}
    
    \caption{Example of a random sample in the different datasets.}
    \label{fig:datasets}
\end{figure}

In table \ref{tab:datasets} we give the different properties of each datasets. Note that an interval in this table means that the quantity in question is uniformly sampled over this interval. For all the samples inlet velocity in all datasets, we sample uniformly between 10 and 50 meters per second which correspond to a Reynolds number\footnotemark[\value{footnote}] from $10^6$ to $5\cdot10^6$ (the characteristic length of our airfoils are 1 meter and the kinematic viscosity\footnotemark[\value{footnote}] of air is roughly $10^{-5}$). The training, validation and the first test datasets are taken from the same distribution, the difference between the second test set (that we call "Test noise" in the table) is the addition of noise when generating the geometries leading to less smooth airfoil. The third and fourth test sets expend the angle of attack and aspect ratio interval respectively. Figure \ref{fig:datasets} shows different geometries with their CFD mesh.

\begin{table}[ht]
    \centering
    \begin{tabular}{|c|c|c|c|c|c|c|}
         \hline
         & Train & Val & Test & Test noise & Test rot & Test big \\
         \hline
        \#Sample & 180 & 20 & 30 & 30 & 30 & 30 \\
        \hline
        Mean \#nodes/sample & 13012 & 12694 & 13519 & 13581 & 13383 & 20499 \\
        \hline
        Min. \#nodes/sample & 9585 & 10028 & 10225 & 10212 & 10137 & 16645 \\
        \hline
        Max. \#nodes/sample & 17102 & 15587 & 16193 & 16448 & 16744 & 24022 \\
        \hline
        Inlet velocity & $[10, 50]$ & $[10, 50]$ & $[10, 50]$ & $[10, 50]$ & $[10, 50]$ & $[10, 50]$ \\
        \hline
        Angle of attack & $[-0.3, 0.3]$ & $[-0.3, 0.3]$ & $[-0.3, 0.3]$ & $[-0.3, 0.3]$ & $[-0.9, 0.9]$ & $[-0.3, 0.3]$ \\
        \hline
        Aspect ratio & $[0.6, 0.8]$ & $[0.6, 0.8]$ & $[0.6, 0.8]$ & $[0.6, 0.8]$ & $[0.6, 0.8]$ & $[0.8, 1]$ \\
        \hline
    \end{tabular}
    \caption{Properties of the different datasets.}
    \label{tab:datasets}
\end{table}

\subsubsection*{Preprocessing}

If we want our networks to learn correctly, we need to normalize our inputs and our outputs. For the inputs, in the training set, we take the mean value $\mu_k$ and the standard deviation $\sigma_k$, of each component $k\in\{0, 1, 2, 3\}$ of our inputs, of all of the nodes in all of the samples and we normalize each component $x_k$ via:
\begin{align*}
    x^{norm}_k = \frac{x_k - \mu_k}{\sigma_k + 10^{-8}}
\end{align*}
where the factor $10^{-8}$ is here for numerical stability.

For the outputs, we would like to do the same thing. However, we remark a particularity in the distribution of the turbulent viscosity at the surface compared to it in the volume. The values of the turbulent viscosity at the surface are of, at least, one order of magnitude smaller and we decide to normalize it independently of the volume values. This greatly improve the learning process at the surface for the turbulent viscosity. Let us write exactly the transformation we do to the outputs. If we keep the notation we used for the inputs normalization and we denote by $y_k$ the $k$-component of the output $y$, we first normalize $y$ via:
\begin{align*}
    y^{norm}_k = \frac{y_k - \mu_k}{\sigma_k + 10^{-8}}
\end{align*}
where $\mu_k$ is the mean value of the k-component of all of the outputs and $\sigma_k$ their standard deviation. In addition to that, for the turbulent viscosity component (\emph{i.e.} for $k = 4$), only for the nodes at the surface of airfoils, we afterward do the transformation:
\begin{align*}
    y^{snorm}_4 &= \frac{y^{norm}_4(\sigma_4 + 10^{-8}) + \mu_4 - \mu^{surf}_4}{\sigma^{surf}_4 + 10^{-8}} \\
    &= \frac{y^{surf}_4 - \mu^{surf}_4}{\sigma^{surf}_4 + 10^{-8}}
\end{align*}
where $\mu^{surf}_4$ is the mean value of the turbulent viscosity of all the surface nodes of all the samples in the training set and $\sigma^{surf}_4$ their standard deviation.

Before we dig further into the task, let us recall the basics of fluid dynamics in order to better understand our problem.

\footnotetext{See section \ref{hydro} for more details.}

\section{Fluid Dynamics overview} \label{hydro}

\emph{This section is completely inspired by the hydrodynamics undergraduate course of Marc Rabeau \cite{courshydro} and the pseudo-book of Tobias Holzmann \cite{mathopenfoam}}

\subsection{Fluid particle}
In a solid, a particle (which can mean an atom, or a bunch of atoms...) is well-located and keeps its neighbours, there is no "movement" inside the solid itself. For a fluid (liquid or gas) it is different, each atom or molecule of fluid has its own velocity that is close to a Brownian motion and the entire medium is deformable. The field of fluid dynamics studies the movement of so-called \emph{fluid particle} that are a mass of atoms or molecules that is big enough to define statistically a significant mean velocity but small enough to consider that its volume tends to zero. The predictions of fluid dynamics only hold when such object exists (which is the case in most of the common applications on earth). Hence, we will call a fluid a continuous medium of such fluid particles where the velocity is deterministically defined at each position. Moreover, every fields defined on a fluid will be assumed smooth.

\subsection{Local and total derivatives}
Let us define a scalar field $T$ depending on the position $r = (x, y, z)$ and on the time $t$, we can write:
\begin{align*}
    dT(r, t) &= \partial_t T(r, t)dt + \partial_1 T(r, t)dx + \partial_2 T(r, t)dy + \partial_3 T(r, t)dz \\ 
    &= \partial_t T(r, t) + \nabla T(r, t) \cdot dl
\end{align*}
where $\partial_i$ is the partial derivative with respect to the $i^{th}$ variable and $\partial_t$ is the partial derivative with respect to time. Moreover, $\nabla T$ is the gradient of the scalar field $T$ and $dl$  is equal to $(dx, dy, dz)$. \\
in the case where $r$ depends on time, by the chain rule, we have:
\begin{align*}
    dT(r(t), t) &= \Big(\partial_t T(r(t), t) + \partial_1 T(r(t), t)\Dot{x}(t) + \partial_2 T(r(t), t)\Dot{y}(t) + \partial_3 T(r(t), t)\Dot{z}(t)\Big)dt \\
    &= \Big(\partial_t T(r(t), t) + \partial_1 T(r(t), t)v_x(t) + \partial_2 T(r(t), t)v_y(t) + \partial_3 T(r(t), t)v_z(t)\Big)dt \\ 
    &= \Big(\partial_t T(r(t), t) + v\cdot \nabla T(r(t), t)\Big)dt
\end{align*}
where, $\Dot{x}$ is the time derivative of $x$, $v$ is the velocity on the trajectory $r$ and $v_i$ its components.

Hence, we define the total derivative $d_t T$ that quantifies the rate of change of the quantity $T$ per unit of time as:
\begin{align}
    d_t T = \partial_t T + v\cdot\nabla T
\end{align}
for all trajectory $r$ and time $t$.

We can easily extend this computation to a vector field $A$ by applying it to each of its component, we write for the $i^{th}$ component:
\begin{align*}
    d_t A_i = \partial_t A_i + v\cdot\nabla A_i
\end{align*}
and we condensate it by writing:
\begin{align}
    d_t A &= \partial_t A + (v\cdot \nabla)A
\end{align}

On the right hand side, the first term in those equations represents the time-variation of the field at the point $r(t)$ and the second term is the variation of the field due to the fact that we are moving along the trajectory $r$, we call this a \emph{convective} term. A lot of complexity in the equations of fluid mechanics arises from this convective term.

\subsection{Transport theorem}

We call a finite volume of fluid particles a volume of control, its closed surface a surface of control and we denote it respectively by $VC$ and $SC$. Such volume can be mobile (in which case the volume of control is a function of time).

In one dimension, a volume of control is given by an interval $[a(t), b(t)]$ and we can look at the variation of the integration of a scalar field $f$ over this volume of control:
\begin{align*}
    d\int_{[a(t), b(t)]} f(x, t)dx = \left(\int_{[a(t), b(t)]} \partial_t f(x,t) dx + f(b(t), t)\Dot{b}(t) - f(a(t), t)\Dot{a}(t)\right)dt
\end{align*}
and this equation can be generalized in three dimension via:
\begin{align}
    \frac{d}{dt}\int_{VC(t)} f(r, t)d\tau = \int_{VC(t)} \partial_t f(r, t)d\tau + \oint_{SC(t)} f(r, t)v(r, t)\cdot n(r, t)dS
\end{align}
where $d\tau$ is an element of volume, $dS$ is an element of surface, $n(r, t)$ is the outward normal of the surface of control at the point $(r, t)$ and $v(r, t)$ is the velocity of the surface of control at the point $(r, t)$. As the surface of control is closed, we can use the particular three dimensional case of the Stokes theorem (namely the Green-Ostrogradsky theorem) and we find:
\begin{align}
    \frac{d}{dt}\int_{VC(t)} f(r, t)d\tau = \int_{VC(t)} \Big(\partial_t f(r, t) + \nabla \cdot (f(r, t)V(r, t))\Big)d\tau
\end{align}
where $\nabla$ is equal to the differential operator $(\partial_1, \partial_2, \partial_3)$

\subsubsection*{Mass conservation}
Applying the precedent theorem to the volumetric mass $\rho$ of the fluid and at a volume of control that follows the stream of the fluid, we find the conservation of the mass in a volume of control. We have, if we denote by $M_{VC(t)}$ the mass included in the volume of control: 
\begin{align*}
    \frac{d}{dt}M_{VC(t)} &= \frac{d}{dt}\int_{VC(t)} \rho(r, t)d\tau \\
    &= \int_{VC(t)} \Big(\partial_t \rho(r, t) + \nabla \cdot (\rho(r, t) v(r, t))\Big)d\tau \\
    &= 0
\end{align*}
where $v$ is the velocity of the fluid on the surface of control. 

This equation is true for all volume of control that follows the stream of the fluid, hence we find that the conservation of the mass of the fluid is ensured by this local equation called \emph{continuity equation}:
\begin{align}
    \partial_t \rho + \nabla\cdot (\rho v) = 0
\end{align}
Moreover, if we are dealing with an incompressible fluid, $\rho$ is constant and the continuity equation can be written as:
\begin{align}
    \nabla\cdot v = 0
\end{align}


\subsection{(Cauchy) Stress tensor}

We call \emph{stress} the force $\sigma(n)$ that is acting (by contact) on a unit surface of normal $n$. For an infinitesimal surface $dS$ which normal $n$ points towards the fluid that is acting on it, the resulting force $df$ can be written as:
\begin{align*}
    df = \sigma(n)dS
\end{align*}

Let us take an infinitesimal cube, we look only at three contiguous faces, and call $\sigma_{ij}$ the force per unit of surface acting on the $i^{th}$ face on the $j^{th}$ direction. We then define the second order \emph{stress tensor} (also known as the \emph{Cauchy stress tensor}) $\sigma$ whose components are $\sigma_{ij}$. By the third law of Newton, we find that, at first order, for an infinitesimal cube, $\sigma(n) = -\sigma(-n)$ which tells us that we only need to know the tensor $\sigma$ to know completely the surface forces acting on the entire cube (and not only on the three contiguous faces chosen previously). Moreover, by an argument on the kinetic moment of this infinitesimal cube, we find that $\sigma$ needs to be symmetric (see \cite{courshydro} p. 39). 

We conclude that for an arbitrary normal $n$, we only need to project the stress tensor on this normal to have the force acting on it, we find:
\begin{align}
    \sigma(n) = \sigma \cdot n = \sigma_{ij}n_j
\end{align}
where we used the Einstein summation convention (summation over repeated indices) on the third expression. \\

\subsubsection*{Viscous stress tensor}
In the case of a fluid with a null velocity field, there is only normal stresses acting on our infinitesimal cube. Moreover, those stresses are isotropic. We call \emph{pressure}, denoted by $p$, the intensity of those stresses. We then find $\sigma = -pI$, where $I$ is the identity matrix of dimension 3. The minus sign is because we took the convention of outward normal when we defined $\sigma(n)$. \\
In the case of a general velocity field, we define the \emph{viscous stress tensor} $\sigma'$ by:
\begin{align}
    \sigma = -pI + \sigma'
\end{align}
We remark that $\sigma'$ is also a second order symmetric tensor.

\subsubsection*{Shear-rate stress tensor}
We can also define the deviatoric part of the stress tensor, we call it the \emph{shear-rate stress tensor} $\sigma_{shear}$ and it is defined by:
\begin{align}
    \sigma_{shear} = \sigma - \frac{1}{3}\Tr(\sigma)I
\end{align}
where $\Tr$ represents the trace operator. 

This tensor can be seen as the shear part of the stress tensor, it only takes in account the tangential forces applied on the surface and get rid of the normal ones. By definition, this tensor is traceless. \\
We will see that in the case of an incompressible fluid, the shear-rate stress tensor and the viscous stress tensor are equal. 

\subsection{Fundamental principle of dynamic}
Let $VC(t)$ be a volume of control that follows the stream of our fluid. In the last section, we defined the local forces acting on the surface of an arbitrary volume of control. We write $g(r, t)$ the total local volumetric forces acting at the point $(r, t)$, one example of such force is the gravitational force. We now have the total forces $F$ acting on $VC(t)$ by:
\begin{align*}
    &F = F_{VC(t)} + F_{SC(t)} \\
    &F_{VC(t)} = \int_{VC(t)} \rho g\, d\tau \\
    &F_{SC(t)} = \oint_{SC(t)} \sigma\cdot n \, dS = \int_{VC(t)} \nabla\cdot\sigma\, d\tau
\end{align*}
where we dropped the argument of the different tensor fields for the sake of simplicity (be aware though that all of those tensor fields, in general, depend on the position $r$, the time $t$ but can also depend on other quantities such as temperature or pressure etc...). In total, we have:
\begin{align}
    F = \int_{VC(t)} \Big(\rho g + \nabla\cdot\sigma\Big)d\tau
\end{align}
We can express the divergence of the stress tensor with the divergence of the viscous stress tensor and the gradient of the pressure:
\begin{align*}
    \nabla \cdot \sigma = -\nabla \cdot (pI) + \nabla\cdot\sigma' = -\nabla p + \nabla\cdot\sigma'
\end{align*}

On the other hand, the variation in time of the quantity of movement is described by:
\begin{align*}
    \frac{d}{dt}\int_{VC(t)} \rho v_i\, d\tau &= \int_{VC(t)} \Big(\partial_t(\rho v_i) + \nabla\cdot(\rho vv_i)\Big) d\tau \\
    &= \int_{VC(t)} \Big(\rho\partial_t v_i + \rho v\nabla\cdot v_i + v_i\big(\partial_t \rho + \nabla\cdot(\rho v)\big)\Big)d\tau \\
    &= \int_{VC(t)} \rho\Big(\partial_t v_i + v\cdot\nabla v_i\Big)d\tau \\
    &= \int_{VC(t)} \rho d_t v_i \, d\tau
\end{align*}
where we used the continuity equation at the third step to simplify the expression. Notice that this equation is still valid for compressible fluid.

In vectorial form, applying the fundamental principle of dynamic, we find:
\begin{align*}
    \frac{d}{dt}\int_{VC(t)} \rho v\, d\tau &= F \\
    \int_{VC(t)} \Big(\rho(\partial_t v + (v\cdot\nabla)v)\Big)d\tau &= \int_{VC(t)} \Big(-\nabla p + \nabla\cdot\sigma' + \rho g\Big)d\tau
\end{align*}
this equation is true for all volume of control that follows the stream of the fluid. Hence, we finally find:

\begin{align}
    \rho\Big(\partial_t v + (v\cdot\nabla)v\Big) = -\nabla p + \nabla\cdot\sigma' + \rho g
\end{align}
Now, we have to give an expression for $\sigma'$ if we want to go further.

\subsection{Inviscid fluids}
Inviscid fluids are fluids where the viscous stress tensor is null. This type of fluid is not realistic and will not be of interest for us in the following but it may be interesting to write the main equations of this problem. 

The fundamental principle of dynamic is written in the following way in this case:
\begin{align}
    \partial_t v + (v\cdot\nabla)v = -\frac{1}{\rho}\nabla p + g
\end{align}
This equation is called \emph{Euler's equation}. Another useful equation can be derived in the steady-state regime after a bit of manipulation. The term on the left hand side can be written as:

\begin{align*}
    (v\cdot\nabla)v = \frac{1}{2}\nabla(\|v\|^2) - v \times (\nabla\times v)
\end{align*}
Moreover, if the fluid is incompressible and if the volumetric force are defined as the gradient of a potential $\Phi$ ($g = -\nabla\Phi$), we find:
\begin{align}
    \nabla\left(\frac{1}{2}\|v\|^2 + \frac{1}{\rho}p + \Phi\right) = v\times(\nabla\times v)
\end{align}
We remark that on a \emph{streamline} (a flow trajectory of the velocity vector field of the fluid), we have:
\begin{align*}
    \nabla\left(\frac{1}{2}\|v\|^2 + \frac{1}{\rho}p + \Phi\right)\cdot dl = v\times(\nabla\times v)\cdot dl = 0
\end{align*}
which means that on every streamline, we have:
\begin{align*}
    \frac{1}{2}\|v\|^2 + \frac{1}{\rho}p + \Phi = C
\end{align*}
where $C$ is a constant. 

If $g$ is the gravitational acceleration, we have $\Phi = -\rho gz $ and the previous equation becomes:
\begin{align}
    \frac{1}{2}\rho\|v\|^2 + p + \rho g z = C
\end{align}
This equation is known as \emph{Bernoulli's equation}.

\subsection{Viscosity}
The deformation of an infinitesimal volume of control can be quantified thanks to the Jacobian $J$ of the velocity. This Jacobian can be decoupled in a symmetric tensor $S$ and a skew-symmetric tensor $W$ via:
\begin{align*}
    J &= S + W \\
    S &= \frac{1}{2}(J + J^t) \\
    W &= \frac{1}{2}(J - J^t)
\end{align*}
where $J^t$ correspond to the transpose of $J$. 

The tensor $W$ represents the pure rotation of the volume of control and the tensor $S$ represents the compression and dilation of the volume of control with respect to a certain basis (as it is symmetric, it can be diagonalizable in an orthonormal basis). Moreover, we remark that $\nabla\cdot v =  \Tr(J) = \Tr(S)$, hence, if the fluid is incompressible, $J$, $S$ and $W$ are traceless. 

\subsubsection*{Newtonian fluid}
We call a \emph{Newtonian fluid}, a fluid where the viscous stress tensor depends linearly on the pure deformation tensor $S$, \emph{i.e.} it exists a fourth order tensor $A$ such that:
\begin{align*}
    \sigma'_{ij} = A_{ijkl}S_{kl}
\end{align*}
where again, we used the Einstein summation convention. 

If the medium is isotropic, it can be shown that such tensor $A$ of order 4 is totally defined by three quantities $A$, $B$ and $C$ such that:
\begin{align*}
    A_{ijkl} = A\delta_{ik}\delta_{jl} + C\delta_{il}\delta_{jk} + B\delta_{ij}\delta_{kl}
\end{align*}
where $\delta$ is the Kronecker tensor of order 2.
Moreover, this tensor is symmetric over its two first argument and its two last arguments because $\sigma$ and $S$ are symmetric tensors. Hence, we conclude that $A = C$ and we have:
\begin{align*}
    A_{ijkl} = 2A\delta_{ik}\delta_{jl} + B\delta_{ij}\delta_{kl}
\end{align*}
The coefficient $A$ will be denoted by $\mu$ and will represent the dynamic viscosity of the fluid whereas the coefficient $B$ will be denoted by $\zeta - \frac{2}{3}\mu$, where $\zeta$ is called the bulk viscosity. With those notations, we find this expression for $\sigma'$:
\begin{align}
    \sigma' = 2\mu\left(S - \frac{1}{3}\Tr(S)I\right) + \zeta \Tr(S)I
\end{align}
Here, we remark that if the fluid is incompressible, we have $\Tr(S) = 0$, hence $\sigma'$ is traceless and we find that $\Tr(\sigma) = 3p$ which gives:
\begin{align*}
    \sigma' &= \sigma - \frac{1}{3}\Tr(\sigma)I \\
    &= \sigma_{shear}
\end{align*}
this proves that the viscous stress tensor is equal to the shear-rate stress tensor for an incompressible fluid as announced previously. 

This tensor (shear-rate stress or viscous stress), with the prediction of the velocity field and the pressure field, will allow us to quantifies the accuracy of the model tested. As we will look at the Jacobian of the velocity field, it will gives us another metrics for the model.

\subsection{Navier-Stokes equation}

The \emph{Navier-Stokes equation} is simply the fundamental principle of dynamic expressed with the viscous stress tensor for Newtonian fluid. We first compute the divergence of the viscous stress tensor in the case of a Newtonian fluid:
\begin{align*}
    \nabla\cdot\sigma' &= 2\mu\nabla\cdot\left(S - \frac{1}{3}\Tr(S)I\right) + \zeta\nabla\cdot(\Tr(S)I) \\
    &= \mu\Delta v + \mu\nabla(\nabla\cdot v) - \frac{2}{3}\nabla(\nabla\cdot v) + \zeta\nabla(\nabla\cdot v) \\
    &= \mu\Delta v + \left(\zeta + \frac{1}{3}\right)\nabla(\nabla\cdot v)
\end{align*}
where $\Delta$ is the laplacian operator (that acts on a vector by applying the laplacian operator at each of its components), where we used the fact that $\Tr(S) = \nabla\cdot v$ and where we assumed that $\mu$ and $\zeta$ were constants of the fluid (which is often the case for homogeneous fluids). 

We now can derive the final form of the Navier-Stokes equation:
\begin{align}
    \rho\Big(\partial_t v + (v\cdot\nabla)v\Big) = -\nabla p + \rho g + \mu\Delta v + \left(\zeta + \frac{1}{3}\mu\right)\nabla(\nabla\cdot v) 
\end{align}
and its variant for an incompressible fluid:
\begin{align}
    \rho\Big(\partial_t v + (v\cdot\nabla) v\Big) = -\nabla p + \rho g + \mu\Delta v
\end{align}
We often find, in the literature, this form of the Navier-Stokes equation:
\begin{align}
    \partial_t v + (v\cdot\nabla) v = -\frac{1}{\rho}\nabla p + g + \nu\Delta v
\end{align}
where $\nu := \mu/\rho$ is called the \emph{kinematic viscosity}.

\subsubsection*{General form of an advection-diffusion equation}
The Navier-Stokes equation for an incompressible fluid is a particular case of the general form of advection-diffusion equation:
\begin{align}
    \partial_t (\rho\phi) = - \nabla\cdot(\rho v\phi) + \nabla\cdot (D\nabla\phi) + S_\phi
\end{align}
where $\phi$ is a scalar field, $\rho$ the inertial coefficient (\emph{i.e.} the coefficient that quantifies the difficulty to change $\phi$ trough time), $D$ the anisotropic diffusion coefficient (\emph{i.e.} $D$ can depends on $r$, in our case the viscosity plays this role by representing the transport of momentum at the molecular scale), $v$ the velocity of the convective medium, and $S_\phi$ a source term (\emph{i.e.} a term that brings/removes energy). 

If we take $\phi = v_i$ (one component of the fluid velocity), $\rho$ the volumetric mass of the fluid, $D = \mu$ and $S_\phi = -\nabla p + \rho g$. We find:
\begin{align*}
    \rho\partial_t v_i + v_i\partial_t\rho &= - v_i\nabla\cdot(\rho v) - \rho v\cdot\nabla v_i + \mu\Delta v_i -\nabla p + \rho gv \\
    \rho\partial_t v_i + \rho v\cdot\nabla v_i &= -\nabla p + \rho g + \mu\Delta v_i
\end{align*}
where we used the continuity equation to go from the first to the second expression. In vectorial form, we find back the Navier-Stokes equation for an incompressible Newtonian fluid:
\begin{align*}
    \rho\Big(\partial_t v + (v\cdot\nabla)v\Big) = -\nabla p + \rho g + \mu\Delta v
\end{align*}


\subsubsection*{Non-dimensionalization of the Navier-Stokes equation}
Solving the Navier-Stokes equation for a set of parameters $\rho$ and $\nu$ and boundary conditions may actually be equivalent to solving a whole family of equations. To enlighten this phenomena we can work with non-dimensional quantities. Moreover, such formulation will help us to see the importance of each term in the partial differential equation. 

In order to do so, let $T$, $L$, $V$, $P$ be characteristics time scale, length scale, velocity scale, pressure scale (respectively) of the problem and $G$ the magnitude of the acceleration of gravity on earth. We define:
\begin{align*}
    t = T\hat{t} \qquad r = L\hat{r} \qquad v = V\hat{v} \qquad p = P\hat{p} \qquad g = G\hat{g}
\end{align*}
All the quantities with a hat are dimensionless and we can update the incompressible Navier-Stokes equation:
\begin{align*}
    \frac{V}{T}\partial_{\hat{t}} \hat{v} + \frac{V^2}{L}(\hat{v}\cdot\hat{\nabla})\hat{v} = -\frac{P}{\rho L}\hat{\nabla}\hat{p} + G\hat{g} + \frac{\nu V}{L^2}\hat{\Delta}\hat{v}
\end{align*}
Let us take $T$ equal to $L/V$ and $P$ equal to $\rho V^2$, we find:
\begin{align*}
    \frac{V^2}{L}\partial_{\hat{t}} \hat{v} + \frac{V^2}{L}(\hat{v}\cdot\hat{\nabla})\hat{v} = -\frac{V^2}{L}\hat{\nabla}\hat{p} + G\hat{g} + \frac{\nu V}{L^2}\hat{\Delta}\hat{v}
\end{align*}
In total, this gives:
\begin{align}
    \partial_{\hat{t}} \hat{v} + (\hat{v}\cdot\hat{\nabla})\hat{v} &= -\hat{\nabla}\hat{p} + \frac{1}{Fr^2}\hat{g} + \frac{1}{Re}\hat{\Delta}\hat{v} \\
    Re = \frac{VL}{\nu} &\qquad Fr = \frac{V}{\sqrt{GL}}
\end{align}
The dimensionless number $Re$ is called the \emph{Reynold's number} and $Fr$ is called the \emph{Froude's number}. \\
In particular, if we neglect the volumetric forces in the equation we find:
\begin{align}
    \partial_{\hat{t}} \hat{v} + (\hat{v}\cdot\hat{\nabla})\hat{v} &= -\hat{\nabla}\hat{p} + \frac{1}{Re}\hat{\Delta}\hat{v}
\end{align}
This equation only depends on the Reynold's number and two flows with the same Reynold's number will have the same dimensionless solution (if we neglect the volumetric forces). 

Moreover, the Reynold's number can be seen as the ratio of the order of magnitude of the inertial term over the order of magnitude of the viscous term:
\begin{align*}
    Re = \frac{\text{inertial term}}{\text{viscous term}} = \frac{\|(v\cdot\nabla)v\|}{\|\nu\Delta v\|} = \frac{V^2/L}{\nu V/L^2} = \frac{VL}{\nu}
\end{align*}
We then have two particular regimes:
\begin{itemize}
    \item $Re \to 0$, viscous term dominates the flow (we call this a Stokes flow)
    \item $Re \to \infty$, inertial term dominates the flow (the Navier-Stokes equation tends towards the Euler's equation for Inviscid fluids)
\end{itemize}

For the Froude's number, we have:
\begin{align*}
    Fr = \left(\frac{\text{inertial term}}{\text{gravity term}}\right)^{1/2} = \left(\frac{\|(v\cdot\nabla)v\|}{\|g\|}\right)^{1/2} = \left(\frac{V^2/L}{g}\right)^{1/2} = \frac{V}{\sqrt{gL}}
\end{align*}

\subsubsection*{Boundary conditions}
We will only talk about boundary conditions at the interface of of fluid and a solid in this part and we will assume that the solid can not adsorb any particle of the fluid. 

Following this assumption, the normal velocity of the fluid must be the same as the normal velocity of the solid:
\begin{align*}
    v_f\cdot n = v_s\cdot n
\end{align*}
where $v_s$ is the velocity of the solid, $v_f$ the velocity of the fluid, and $n$ the normal pointing in the direction of the fluid. 

For inviscid fluids, there is no condition on the tangential velocity of the fluid. However, for viscous fluids, an experimental boundary condition is the equality of the tangential velocity of the fluid and the solid which gives, with the first boundary condition, the global equality between the fluid and the solid velocity:
\begin{align*}
    v_f = v_s
\end{align*}

\subsection{Turbulence}

Navier-Stokes equations are a complex set of partial differential equations. We often study analytically  Stokes flows (\emph{i.e.} flows with low Reynolds compared to 1) not because those flows are preponderant in nature but because they are easier case to treat. Actually, most of the flows encountered in nature have a very high Reynolds number compared to 1. 

At enough high Reynolds number, new patterns of the motion of the fluid arise. Those patterns looks chaotic and very difficult to study. We call this regime the \emph{turbulence regime} and one way to describe those phenomenons is to describe it statistically.

\subsubsection*{Statistical description of turbulence}
We would like to do a lot of experiments in parallel, look at the velocity field in every points for each of those experiments and do a statistical analysis. Unfortunately, this can not be done in reality and we must do an \emph{ergodic} hypothesis. We will assume that this \emph{ensemble mean} is equal to the mean field velocity in time for a characteristic time $T$ that is bigger to all time scales of fluctuations. However, this characteristic time $T$ needs to be small compare to the time scale of macroscopic changes in the flow. We will call \emph{stationary flow}, a flow where $T$ can be taken as high as we want (\emph{i.e.} there is no macroscopic change of the motion in time). 

We define for all $(r, t)$:
\begin{align*}
    \Bar{v}(r, t) = \frac{1}{T}\int_{t - \frac{T}{2}}^{t + \frac{T}{2}} v(r, s)ds
\end{align*}
and we will write $v = U + u$, where $U = \Bar{v}$. In the same way, we will write $p = P + p'$ where $P = \Bar{p}$ for the pressure and $\rho = \Bar{\rho} + \rho'$ for the volumetric mass. The ensemble mean and the mean in time of the fluctuations terms (such as $u$, $p'$ and $\rho'$) is null.

Injecting this in the continuity equation, we find:
\begin{align*}
    \partial_t (\Bar{\rho} + \rho') + \nabla\cdot(\Bar{\rho} U + \Bar{\rho}u + \rho' U + \rho' u) = 0
\end{align*}
taking the ensemble mean of this equation, we find:
\begin{align}
    \partial_t \Bar{\rho} = \nabla\cdot\left(\Bar{\rho}U + \overline{\rho' u}\right)
\end{align}
For an incompressible fluid, we find:
\begin{align}
    \nabla\cdot U = 0
\end{align}
and as $\nabla\cdot v = \nabla\cdot U + \nabla\cdot u$, we also find $\nabla\cdot u = 0$.  The mean flow and the fluctuation flow are both incompressible flows.

\subsubsection*{Reynolds-Average-Navier-Stokes (RANS)}
Now, let us inject the mean-fluctuation decomposition into the Navier-Stokes equation for incompressible fluids, for clarity, we write everything in coordinates using the Einstein summation convention:
\begin{align*}
    \partial_t (U_i + u_i) + (U_j + u_j)\partial_j(U_i + u_i) = -\frac{1}{\rho}\partial_i(P + p') + g + \nu\partial^2_{jj}(U_i + u_i)
\end{align*}
taking the ensemble mean of this equation, we find:
\begin{align*}
    \partial_t U_i + U_j\partial_j U_i + \overline{u_j\partial_j u_i} &= -\frac{1}{\rho}\partial_i P + g + \nu\partial^2_{jj} U_i \\
    \partial_t U_i + U_j\partial_j U_i &= -\frac{1}{\rho}\partial_i P + g + \nu\partial^2_{jj} U_i - \partial_j\overline{(u_ju_i)}
\end{align*}
where we used $\nabla\cdot u = 0$ in the second expression. 

We define $(\sigma_t)_{ij} := -\rho\overline{u_iu_j}$ a symmetric second order tensor and we call it the \emph{Reynolds stress tensor}. Hence, the last expression can be written as:
\begin{align}
    \rho\Big(\partial_t U + (U\cdot\nabla)U\Big) = -\nabla P + \rho g + \nabla\cdot(\Bar{\sigma}' + \sigma_t)
\end{align}
where $\Bar{\sigma}'$ is the viscous stress tensor for the mean field velocity. \\
This equation is called the \emph{Reynolds-Average-Navier-Stokes (RANS) equation}. It only differs from Navier-Stokes equation by the term $\sigma_t$ that contains the correlation of the fluctuations of the components of the velocity field at a each point $(r, t)$. This term corresponds to the diffusion of quantity of movement through turbulence.

\subsubsection*{Closure problem}
We now have another problem, in the RANS equation, we have five unknown quantities whereas we have only four equations. We need either to link the Reynolds stress tensor to the four other unknowns or to find another equation to close the problem. We describe in the following one attempt of the first strategy that is widely use nowadays.

\subsubsection*{Turbulent viscosity}
The \emph{Boussinesq hypothesis} is an hypothesis that links the Reynolds stress tensor to the symmetric part of the Jacobian of the mean velocity field:
\begin{align}
    \sigma_t &= 2\mu_t\left(\Bar{S} - \frac{1}{3}\Tr(\Bar{S})I\right) - \frac{2}{3}\rho kI \\
    \Bar{S}_{ij} &= \frac{1}{2}(\partial_i U_j + \partial_j U_i)
\end{align}
where $\mu_t$ is called the \emph{turbulent dynamic viscosity (also known as eddy dynamic viscosity)} and $k$ is the kinematic energy of the turbulence per unit of mass, \emph{i.e.}:
\begin{align*}
    k = \frac{1}{2}\overline{u_ju_j}
\end{align*}
This expression describes mathematically the fact that the turbulence phenomena can be seen as a macroscopic diffusion of quantity of movement in analogy to the classical viscosity diffusion. The term involving $k$ is necessary to recover the trace of the Reynolds stress tensor. As the dissipation via turbulence is expected to be large compared to the dissipation via viscosity, we should have $\mu_t$ large compared to $\mu$. 

By defining an effective pressure $p_{eff} := P + 2\rho k$ and an effective viscosity $\nu_{eff} := \nu + \nu_t$ where $\nu_t:= \mu_t/\rho$ is the \emph{turbulent kinematic viscosity}, we have:
\begin{align}
    \partial_t U + (U\cdot\nabla)U = -\frac{1}{\rho}\nabla p_{eff} + g + \nu_{eff}\Delta U
\end{align}
and we recover a Navier-Stokes equation with an effective pressure and an effective viscosity. 

Now, we manage to express the Reynolds stress tensor out of our four previous unknown quantities but by doing that, we defined a new unknown which is the $turbulent viscosity$ $\nu_t$. A model to know this turbulent viscosity is called a \emph{turbulent model}.

\subsubsection*{Turbulent models}
There are plenty of different turbulent models that have been developed. Each model is empirical and have its own field of application. For our purpose, we will use the \emph{Spalart-Allmaras model} \cite{doi:10.2514/6.1992-439} which is a one-equation model for the turbulent viscosity. In order to do so, we define various quantities and constant that are all chosen experimentally. The equation is:
\begin{align}
    \partial_t \Tilde{\nu} + U_j\partial_j \Tilde{\nu} = c_{b1}(1 - f_{t2})\Tilde{S}\Tilde{\nu} - \left(c_{w1}f_w - \frac{c_{b1}}{\kappa^2}f_{t2}\right)\left(\frac{\Tilde{\nu}}{d}\right)^2 + \frac{1}{\sigma}\Big(\partial_j\big(\nu + \Tilde{\nu}\big)\partial_j\Tilde{\nu}\Big) + c_{b2}\partial_i\Tilde{\nu}\partial_i \Tilde{\nu}
\end{align}
with:
\begin{align*}
    \nu_t = \Tilde{\nu}f_{v1}\,, \quad f_{v1} = \frac{\chi^3}{\chi^3 + c^3_{v1}}\,, \quad \chi = \frac{\Tilde{\nu}}{\nu}\,&, \quad \Tilde{S} = \Omega + \frac{\Tilde{\nu}}{\kappa^2d^2}f_{v2}\,, \quad \Omega = \sqrt{2W_{ij}W_{ij}}\,,  \\
    f_{v2} = 1-\frac{\chi}{1 + \chi f_{v1}}\,, \quad f_w = g\left(\frac{1+ c_{w3}^6}{g^6 + c_{w3}^6}\right)^{1/6}\,&, \quad g = r + c_{w2}(r^6 - r)\,, \quad r = \min\left(\frac{\Tilde{\nu}}{\Tilde{S}\kappa^2d^2}, 10\right) \\
    W_{ij} = \frac{1}{2}(\partial_j U_i - \partial_i U_j)\,&, \quad f_{t2} = c_{t3} e^{-c_{t4}\chi^2}
\end{align*}
and for the constants:
\begin{align*}
    c_{b1} = 0.1355\,,\quad \sigma = \frac{2}{3}\,,\quad c_{b2} = 0.622\,&,\quad \kappa = 0.41 \,,\quad c_{w2} = 0.3\,,\quad c_{w3} = 2\,,\\
    c_{v1} = 7.1\,,\quad c_{t3} = 1.2\,,\quad c_{t_4} = 0.5\,&,\quad c_{w1} = \frac{c_{b1}}{\kappa^2} + \frac{1 + c_{b2}}{\sigma} 
\end{align*}
This allows us to close the problem and move forward to numerical resolution. Now let us review some basic results of the theory of partial differential equations in order to gain a priori knowledge on Navier Stokes equations.

\section{Partial Differential Equations} \label{PDE}

\emph{This section is completely inspired by Lawrence C. Evans' book "Partial Differential Equation" \cite{evans}} 

Partial Differential Equations (PDE) are equations that govern loads of differents phenomenons in nature. Their studies is a hard task as they describe very differents behaviours. Actually, even though the Cauchy-Kowalevsky theorem gives the existence of solutions for a wide class of PDE, there is no hope to find a generic framework for solving all PDE. 

What is often done in the study of PDE is to separate the linear from the non-linear ones. We will do a very short comment on both in the next two subsections by giving some classification and an example of resolution that will be important for us later. 

Moreover, studying PDE means looking at the regularity of the solutions built. The theory of PDE introduces weak form of regularity that we will not discuss here. Every solutions will be considered enough smooth for the problem we are discussing.

\subsection{Linear PDE}
First, we call a PDE \emph{linear} when it can be written as a sum of partial derivatives (even of order 0) with coefficient that depends only on the points we are looking at (the coefficients do not depend on the unknown or its partial derivatives).

We will present three archetypal PDE whose form can be extended to a generic classification. In the following, we will call $U$ an open set of $\mathbb{R}^n$.
\subsubsection*{Laplace's equation}
The \emph{Laplace's equation} is an equation that is written under the form:
\begin{align*}
	-\Delta u = 0 \quad in \; U
\end{align*}
and the \emph{Poisson's equation} is the Laplace's equation with a source term:
\begin{align*}
	-\Delta u = f \quad in \; U
\end{align*}
A boundary term can be added to possibly have uniqueness of a solution. 

Those famous equations are the archetype of the \emph{second order linear elliptic equation} class of linear PDE, this is why we kept the minus sign in their definitions in order to stick with the definition of an elliptic PDE.

\subsubsection*{Heat equation}
The \emph{heat equation} is also a common equation but this one involves time in its definition:
\begin{align*}
	\partial_t u -\Delta u = 0 \quad in \; U\times (0, +\infty)
\end{align*}
and its \emph{non homogeneous} form:
\begin{align*}
	\partial_t u -\Delta u = f \quad in \; U\times (0, +\infty)
\end{align*}
As previously, a boundary term can be added for uniqueness properties. Those equations are the archetype of the \emph{second order linear parabolic equation} class of PDE. The difference between this type of equations and the previous ones is that we have a evolution in time in this case, actually, a second order linear parabolic equation is a second order linear elliptic equation with a first order time-derivation term added.

\subsubsection*{Wave equation}
The \emph{wave equation} appears, as the previous cases, naturally in physics and is of second order in time:
\begin{align*}
	\partial^2_{tt} u -\Delta u = 0 \quad in \; U\times (0, +\infty)
\end{align*}
and its \emph{non homogeneous} form:
\begin{align*}
	\partial^2_{tt} u -\Delta u = f \quad in \; U\times (0, +\infty)
\end{align*}
As previously, a boundary term can be added for uniqueness properties. Those equations are the archetype of the \emph{second order linear hyperbolic equation} class of PDE. Actually, a second order linear hyperbolic equation is a second order linear elliptic equation with a second order time-derivation term added. 

In our case, as we are studying a steady state problem, we will focus on Laplace and Poisson's equations in dimension greater or equal to 2 (as the resolution is immediate in one-dimension). The good news with those equations is that we can explicitly express the solutions of the equations.

\subsubsection*{Fundamental solution of Laplace's equation}
First, we show the rotation invariance of the Laplace's equation when $U = \mathbb{R}^n$. Let $O \in \mathrm{O}(n)$ be an orthogonal matrix acting on $\mathbb{R}^n$, $u\in C^2(\mathbb{R}^n)$ and $x\in\mathbb{R}^n$, we write $v := u\circ O$:
\begin{align*}
	\Delta v(x) &= \nabla \cdot\nabla(u\circ O)(x) \\
	&= \nabla \cdot O^t \nabla u(Ox) \\
	&= O_{ji} \partial_i(\partial_j u \circ O)(x) \\
	&= O_{ji}O_{ki} (\partial^2_{kj} u)(Ox) \\
	&= \delta_{kj}(\partial^2_{kj} u)(Gx) \\
	&= \Delta u(Ox)
\end{align*}
where we used $O^tO = I_n$ as $O$ is orthogonal. Hence, if $u$ satisfies the Laplace's equation over $\mathbb{R}^n$, $v$ satisfies it too. 

This symmetry incites us to look for a radial solution. Let us define $r(x) := \sqrt{\sum_{i = 1}^n x^2_i}$ and compute the laplacian of a smooth function $v$ over $\mathbb{R}^{\times}_+$ depending only on $r$, :
\begin{align*}
	\Delta (v\circ r)(x) &= \nabla\cdot\nabla(v\circ r)(x) \\
	&= \nabla\cdot \left (v'\circ r(x)\nabla r(x) \right ) \\
	&= \nabla r(x)\cdot \nabla (v'\circ r)(x) + (v'\circ r)(x)\Delta r(x)  \\
	&= v''\circ r(x)\|\nabla r(x)\|^2 + v'\circ r(x)\Delta r(x) \\
	&= v''\circ r(x) + \frac{n - 1}{r(x)}v'\circ r(x) \\
	&=  v''(r(x)) + \frac{n - 1}{r(x)}v'(r(x))
\end{align*}
Hence, as $r$ is surjective from $\mathbb{R}^n$ to $\mathbb{R}^{\times}_+$, the Laplace's equation reduces to the Ordinnary Differential Equation (ODE):
\begin{align*}
	v''(r) + \frac{n - 1}{r}v'(r) = 0\, , \quad r\in\mathbb{R}^{\times}_+
\end{align*}
If $v' \neq 0$ (otherwise the solution is constant), an immediate solution of this ODE is:
\begin{align*}
	v'(r) = \frac{a}{r^{n-1}}\, ,\quad a\in\mathbb{R} 
\end{align*}
which gives:
\begin{align*}
	v(r) = \begin{cases}
				a\log r + c \quad &\text{if}\; n = 2 \\
				-\frac{a}{(n-2)r^{n-2}} + c \quad &\text{if}\; n > 2
			\end{cases}
\end{align*}
with $c\in\mathbb{R}$. 

We call the \emph{fundamental solution} of Laplace's equation the function defined over $\mathbb{R}^n\setminus \{0\}$:
\begin{align*}
	\Phi(x) = \begin{cases}
				-\frac{1}{2\pi}\log |x| \quad &\text{if}\; n = 2 \\
				\frac{1}{n(n-2)\alpha(n)}\frac{1}{|x|^{n-2}} \quad &\text{if}\; n > 2
			\end{cases}
\end{align*}
where $\alpha(n)$ is the volume of the unit ball in $\mathbb{R}^n$.

\subsubsection*{Non homogeneous solution}
Now tat we have a solution for the homogeneous problem, we look at the non homogeneous problem. 
Let $f\in \mathcal{C}^2_c(\mathbb{R}^n)$ a twice differentiable function with compact support. We define $u$ such that, for all $x\in\mathbb{R}^n$:
\begin{align*}
	u(x) &:= \int_{\mathbb{R}^n} \Phi(x - y)f(y)dy \\
	&= \int_{\mathbb{R}^n} \Phi(y)f(x - y)dy
\end{align*}
which is well defined as $\Phi$ is always integrable around 0 and $f$ has a compact support. We are going to show that $u$ is a solution of the Poisson's equation. As $f\in \mathcal{C}^2_c(\mathbb{R}^n)$, it is straightforward to show that $u\in\mathcal{C}^2(\mathbb{R}^n)$, we write:
\begin{align*}
	\Delta u(x) &= \int_{\mathbb{R}^n} \Phi(y)\Delta f(x - y)dy \\
	&= \int_{B_{\epsilon}} \Phi(y)\Delta f(x - y)dy + \int_{B_{\epsilon}^c} \Phi(y)\Delta f(x - y)dy \\
	&:= I_{\epsilon} + J_{\epsilon}
\end{align*}
where $B_{\epsilon}$ is the open ball of center 0 and radius $\epsilon$ of $\mathbb{R}^n$. We have:
\begin{align*}
	I_{\epsilon} &\leqslant C\|\nabla^2f\|_{L^{\infty}(\mathbb{R}^n)}\int_{B_{\epsilon}} |\Phi(y)|dy \\
	&\leqslant \begin{cases}
					C\epsilon^2|\log\epsilon| \quad &\text{if}\; n=2 \\
					C\epsilon^2 \quad &\text{if}\; n > 2
				\end{cases}
\end{align*}
for $\epsilon \leqslant e^{-1}$, where $\nabla^2f$ is the Hessian of $f$ and $C$ denotes a constant (that is not necessarily the same from an equality to another). We also have:
\begin{align*}
	J_{\epsilon} &= \int_{B_{\epsilon}^c} \Phi(y)\Delta_x f(x - y)dy \\
	&= \int_{B_{\epsilon}^c} \Phi(y)\Delta_y f(x - y)dy \\
	&= -\int_{B_{\epsilon}^c} \nabla\Phi(y)\cdot\nabla_y f(x-y)dy + \int_{\partial B_{\epsilon}} \Phi(y) \nabla_y f(x - y) \cdot n(y) dS(y) \\
	&:= K_{\epsilon} + L_{\epsilon}
\end{align*}
where $n(y)$ is the outward unit normal and $dS(y)$ an element of surface of $\partial B_{\epsilon}$ at the point $y$. For $L_{\epsilon}$, we have:
\begin{align*}
	L_{\epsilon} &\leqslant C\|\nabla f\|_{L^{\infty}(\mathbb{R}^n)} \int_{\partial B_{\epsilon}} |\Phi(y)|dS(y) \\
	&\leqslant \begin{cases}
					C\epsilon|\log \epsilon| \quad &\text{if}\; n=2 \\
					C\epsilon \quad &\text{if}\; n > 2
				\end{cases}
\end{align*}
And for $K_{\epsilon}$:
\begin{align*}
	K_{\epsilon} &= -\int_{B_{\epsilon}^c} \nabla\Phi(y)\cdot\nabla_y f(x-y)dy \\
	&= \int_{B_{\epsilon}^c} \Delta\Phi(y)f(x-y)dy - \int_{\partial B_{\epsilon}} f(x - y)\nabla \Phi(y)\cdot n(y)dS(y) \\
	&= -\int_{\partial B_{\epsilon}} f(x - y)\nabla \Phi(y)\cdot n(y)dS(y) \\
	&= - \frac{1}{n\alpha(n)\epsilon^{n - 1}}\int_{\partial B(x, \epsilon)} f(y)dS(y)
\end{align*}
where we used the fact that $n(y) = -\frac{y}{\epsilon}$ and $\nabla \Phi(y) = -\frac{y}{n\alpha(n)\epsilon^{n}}$ when $y\in\partial(B_{\epsilon}^c)$ and where $B(x, \epsilon)$ is the open ball of center $x$ and radius $\epsilon$. Moreover, we have that $n\alpha(n)\epsilon^{n-1}$ is the "volume" of $\partial B(x, \epsilon)$ which means that $K_{\epsilon}$ is the opposite of the mean value of $f$ over $\partial B(x, \epsilon)$. In particular, we have:
\begin{align*}
	K_{\epsilon} \to -f(x) \quad \text{as} \quad \epsilon\to 0
\end{align*}
In total, by sending $\epsilon$ to 0, we find, for all $x\in\mathbb{R}^n$:
\begin{align*}
	-\Delta u(x) = f(x)
\end{align*}
which means that $u$ is a solution of the Poisson's equation. 

Unfortunately, Navier-Stokes' equation is not a linear PDE which means that we need new tools to try to solve it.

\subsection{Non-Linear PDE}
In the non-linear settings, there is no such partial classification as for linear PDE. 

One way to treat our problem is to express it, when it is possible, in a variational way. A functional is defined over a well chosen space of functions (typically a Sobolev space) and the solution of the PDE we are looking at is a minimum or a critical point of this functional. Under some regularities of the functional (convexity, coercivity...) we can show that such minimum exists. This technique can, for example, be applied to the Euler's equation that we saw in the review of fluid dynamics. However, it is not straightforward to express dissipation in such variational techniques and this induces that Navier-Stokes' equation does not fall into this category of problem. 

Other techniques such as fix point formulation or viscosity solution can be used to treat other non-linear PDE but here again, Navier-Stokes' equation, in its general form, is not solvable via those methods. 

Another way to treat non-linear PDE is to try to linearize the problem by finding a mapping that transforms our non-linear PDE to a linear one. To illustrate this method, we are going to present the \emph{Hopf-Cole transformation} by an example. Let us define the PDE:
\begin{align*}
	\partial_t u - a\Delta u + b\|\nabla u\|^2 = 0 \quad \text{in}\; \mathbb{R}^n\times(0, +\infty)	
\end{align*}
and let $\phi$ be a function from $\mathbb{R}$ to $\mathbb{R}$. We try the change of variable $w := \phi(u)$, we have:
\begin{align*}
	\partial_t w &= \phi'(u)\partial_t u \\
	&= \phi'(u)\left (a\Delta u - b\|\nabla u\|^2\right ) \\
	&= a\Delta w - \left (a\phi''(u) + b\phi'(u)\right )\|\nabla u\|^2 
\end{align*}
Hence, if $\phi'$ satisfies the ordinary differntial equation (ODE) $a\phi'' + b\phi' = 0$, we have:
\begin{align*}
	\partial_t w - a\Delta w = 0
\end{align*}
which is of the form of the heat equation. Actually, $\phi(u) := e^{-\frac{b}{a}u}$ solves the previous ODE. For this PDE, the transformation $w := e^{-\frac{b}{a}u}$ is called the Hopf-Cole transformation and allow us to solve the problem by solving the heat equation. The \emph{Koopman theory} \cite{koopman} goes in that direction too and roughly tells us that every non linear PDE can be reformulated in a linear PDE with the help of an infinite dimensional representation. 

All this materials will help for the task of approximating the solutions of the RANS equation. Let us now dig into the resolution of the main problem.

\section{Proposed work} \label{experiments}

We recall the task, from a domain, a geometry and conditions at the boundaries of our domain, we would like to find the steady-state solution of the incompressible RANS equations. Our inputs are CFD meshes that can be viewed as unstructured point clouds or graphs but not as images. Hence, we cannot use the framework of Convolutionnal Neural Networks to tackle our problem. Moreover, we would like to use the physic equations and the theory of PDE as a strong a priori. 

The incompressible steady-state RANS equations can be written as (neglecting the source term):
\begin{align*}
(U\cdot \nabla)U &= -\frac{1}{\rho}\nabla p + (\nu + \nu_t) \Delta U \\
\nabla\cdot U &= 0
\end{align*}

where $U$ is the time-average velocity, $\rho$ the density, $p$ an effective pressure, $\nu$ the kinematic viscosity and $\nu_t$ the kinematic turbulent viscosity, see Sec ... To this, we have to add the Spallart-Allmaras equation in order to close the problem plus some boundary constraints. Let us assume that a solution exists for the incompressible steady-state RANS equations given constraints at the boundaries of our domain and let us call $Z := (U_x,\, U_y,\, p,\, \nu_t)$ such solution. If we represent our domain and the constraints at the boundary of this domain as a mapping $g$ defined over $\mathbb{R}^2$, we can express our task as finding an approximation of a mapping $\mathcal{F} : g\mapsto Z(g)$ between two functional spaces. In our task, those functions $g$ and $Z$ are discretized over a CFD mesh used for the simulation of the data. 

For the metrics, we will use the \emph{Mean Square Error} (MSE) over the surface of the airfoil plus the MSE over the volume of the domain for training, \emph{i.e.} our loss $\mathcal{L}$ will be written as:
\begin{align*}
    \mathcal{L} &:= \frac{1}{|\mathcal{V}|}\sum_{x\in \mathcal{V}} \|f_\theta(x) - y\|_2^2 + \frac{\lambda}{|\mathcal{S}|}\sum_{x\in \mathcal{S}} \|f_\theta(x) - y\|_2^2 \\
    &:= \mathcal{L}_\mathcal{V} + \lambda\mathcal{L}_\mathcal{S}
\end{align*}
where $\lambda \in\mathbb{R}$, $x$ are the inputs, $y$ the targets, $\mathcal{V}$ the set of points that are not at the surface of the airfoil, $\mathcal{S}$ the set of points that are at the surface of the airfoil, $\|\cdot\|_2$ the Euclidean norm and $f_\theta$ our network. The coefficient $\lambda$ allows us to balance the weight of the error at the surface of the geometry and over the volume, we will set it to 1 in the following.

We will also compute, a posteriori, at the surface of the geometry, the wall shear stress $W$ (see section \ref{hydro}) and the isostatic pressure $I$ (which is the unit outward normals times the opposite of pressure field of the surface). This allows us to recover the stress forces at the surface of the geometry and the \emph{global coefficients} of the geometry, which are the \emph{lift} $L$ and the \emph{drag} $D$ defined as:
\begin{align*}
L &:= \left (\oint_{S_{g}} \sigma\cdot n\, dS\right )_x \\
D &:= \left (\oint_{S_{g}} \sigma\cdot n \,dS\right )_y
\end{align*}
where $S_g$ is the geometry surface (in 1D here, \emph{i.e.} a contour), $\sigma$ the Cauchy Stress Tensor and $n$ the outward unit normal of the surface. By injecting the definition of the Cauchy stress tensor into those definitions, we find, by calling $F$ the vector $(L, D)$:
\begin{align*}
F &= -\oint_{S_g} pn\,dS + \oint_{S_g} \sigma_{shear}\cdot n \,dS \\
&= \oint_{S_g} I\,dS + \oint_{S_g} W\,dS \\
I_i &:= pn_i \\ 
W_{ij} &:= (\sigma_{shear})_{ij}n_j = 2(\nu + \nu_t)\left (S - \frac{1}{3}\Tr (S) I\right )_{ij}n_j \\
S_{ij} &:= \frac{1}{2}(\partial_i U_j + \partial_j U_i) 
\end{align*}

The derivatives of the velocity field will be done by Finite Volume Element method with the help of PyVista.

Hence, the task will be to have the minimal error on both those global coefficients and the volumetric MSE. 

Finally, we underlined three important difficulties we face in this work:

\textbf{Numerical complexity.} As we already mentioned it, the goal of this task is to validate a model in order to extend it to a 3D case. Although the number of nodes of each CFD mesh in 2D may be still reasonable for nowadays hardware, it is clearly impossible to train a model on a 3D CFD mesh by keeping every single nodes. Downsampling or other techniques must be used. Hence, the model has to be robust to those techniques.

\textbf{Long-range impact.} Even though the equations are locals, we have seen previously that the solution at a point may be impacted by long-range interaction between nodes. This means that it would certainly be better to make the information flows across all range in our model. In classical GNN, we could for example stacks several layers but we know that this goes with difficulties in the optimization process. Another techniques would be to get rid of the CFD mesh and to create a fully connected graph over our point cloud, but this goes against the reduction of numerical complexity. Our model needs to take this in account to.

\textbf{Small quantity of data.} CFD simulations are costly, and again, even though we could have access to potentially infinite data in our toy dataset case, it would not be representative of the 3D case. The dataset must be small. However, the good news is that the inference may be done on geometries and boundary conditions that are pretty close to the one we find in the training set.

\subsection{Baselines}


Here, we present the different results for our baselines and why we chose them. We will test it on two different contexts to allow it to be in the same condition as our architecture. For each task, we have in input the position of each nodes, the velocity of the flow at infinity and an information of the geometry through the signed distance function.

For each of the models, we will chose randomly an airfoil in the test set and display ground truth versus predicted fields. We will also show the global coefficients curves over all the test set and the \emph{Mean Average Error} (MAE) of it in a table in addition to the MSE over the volume and the surface. Moreover, we will add the scores of the other (harder) test sets in this table.
 
\subsubsection*{GraphSAGE}
The GraphSAGE layer \cite{SAGE} is the basic inductive type layer, it will give us a first indicator of the difficulty of the task. We used a 4-layers GraphSAGE network with 64 channels and ReLU activation. The optimization is done with Adam, a one-cycle cosine learning rate of maximum $10^{-3}$, a batch size of 1 and during 1000 epochs.

As already stated, we trained the model over two different settings, the first one is to train the model with all the nodes of each sample and to take the CFD mesh as input graph. In this setting, the inference (for validation or test) is also done with all the nodes and with the CFD mesh. In the second setting, the training is done by sampling 1600 different nodes for each sample and for each epoch of training and take as input graph the graph that connects all the nodes that are closer than a radius 0.1 from each other in Euclidian distance. This second setting is to take in account the numerical complexity difficulty that arise in the 3D case. The value of 1600 for the sampling is pretty arbitrary (from 10 to 20\% of the total amount of nodes per sample) but the value of 0.1 for the radius is taken in such way that the radius graph holds in memory. The value of the radius has to be compared with the size of the normalized domain which goes from -4 to 4 in the $x$ and $y$-direction which means that we are looking at very local dependence between nodes.

The scores of this model are given in table \ref{GSAGE_score}, $\mathcal{L}_\mathcal{V}$, $\mathcal{L}_\mathcal{S}$ and \emph{glob. MAE} hold for the MSE over the volume, the surface and the global coefficients MAE respectively, \emph{x-WSS}, \emph{y-WSS}, \emph{x-IsoP} and \emph{y-IsoP} hold for the components of the Wall Shear Stress and the Isostatic Pressure respectively. Moreover, the MSE is given in terms of the normalized quantities whereas the global coefficients are given in terms of the unnormalized quantities. We remark that learning over fewer nodes but by controlling the distance at which each nodes communicates with others gives better results (over all nodes) in the test setting compare to the full nodes training over the CFD mesh.


\begin{table}
    \centering
    \begin{tabular}{|c|c|c|c|c|c|c|c|}
        \hline
        \multicolumn{3}{|c|}{} & Val & Test & Test noise & Test rot & Test big \\
        \hline
        \multirow{6}{*}{CFD Mesh} & \multicolumn{2}{c|}{$\mathcal{L}_\mathcal{V}$} & 0.605 & 0.361 & 0.370 & 1.86 & 0.927 \\
         & \multicolumn{2}{c|}{$\mathcal{L}_\mathcal{S}$} & 0.526 & 0.304 & 0.327 & 1.23 & 0.660 \\
         \cline{2-8}
         & \multirow{4}{*}{glob. MAE} & x-WSS & 0.075 & 0.165 & 0.154 & 0.495 & 0.212 \\
         & & y-WSS & 0.122 & 0.101 & 0.099 & 0.229 & 0.274 \\
         & & x-IsoP & 35.7 & 19.9 & 20.7 & 140 & 30.7 \\
         & & y-IsoP & 126 & 93.4 & 95.2 & 219 & 127 \\
         \hline
         \multirow{6}{*}{Sampled radius graph} & \multicolumn{2}{c|}{$\mathcal{L}_\mathcal{V}$} & 0.353 & 0.178 & 0.188 & 1.54 & 0.521 \\        
         & \multicolumn{2}{c|}{$\mathcal{L}_\mathcal{S}$} & 0.398 & 0.225 & 0.248 & 1.01 & 0.588 \\
         \cline{2-8}
         & \multirow{4}{*}{glob. MAE} & x-WSS & 0.054 & 0.117 & 0.132 & 0.481 & 0.144 \\
         & & y-WSS & 0.082 & 0.075 & 0.074 & 0.187 & 0.197 \\
         & & x-IsoP & 19.9 & 10.5 & 11.2 & 65.3 & 43.8 \\
         & & y-IsoP & 73.8 & 57.9 & 58.7 & 121.9 & 86.1 \\
         \hline
    \end{tabular}
    \caption{\label{GSAGE_score} Scores of the GraphSAGE model.}
\end{table}

\begin{figure}
    \begin{subfigure}{\textwidth}
      \centering
      \caption{Ground truth}
      \includegraphics[width=\linewidth]{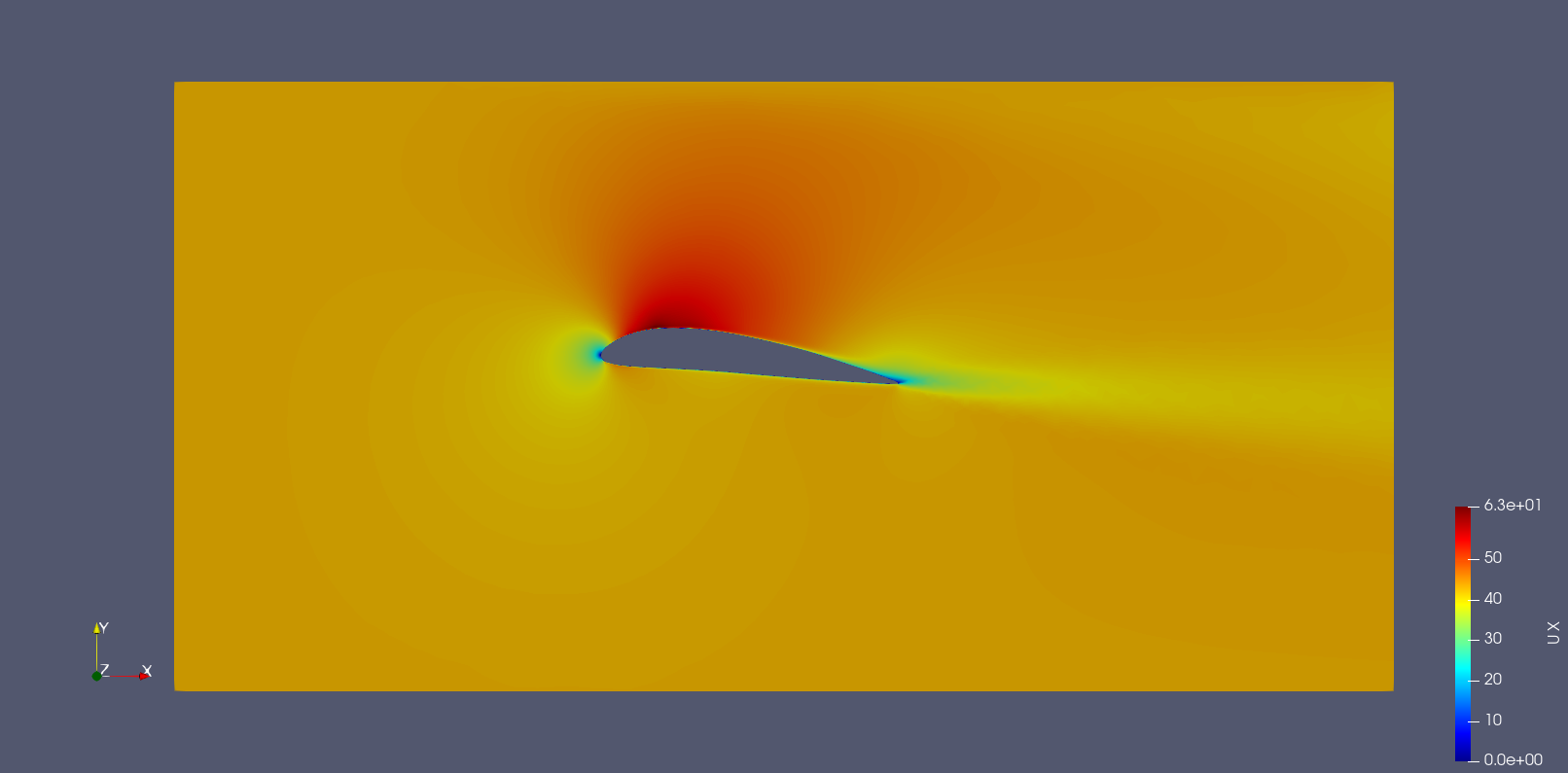}  
      \label{fig:SAGE_true}
    \end{subfigure}

    \begin{subfigure}{.5\textwidth}
      \centering
      \includegraphics[width=\linewidth]{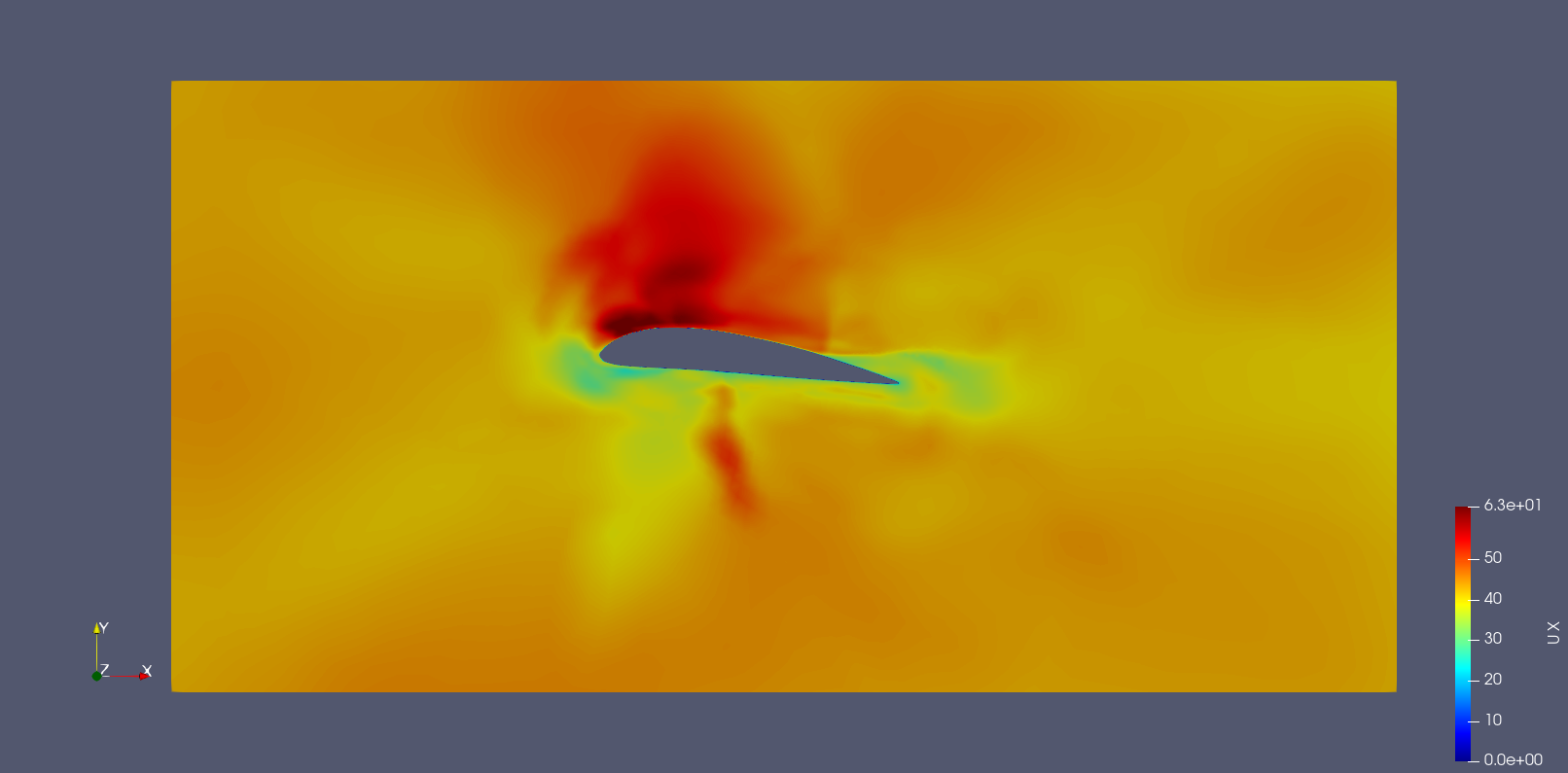}  
      \caption{CFD mesh}
      \label{fig:SAGE_CFD}
    \end{subfigure}
    \begin{subfigure}{.5\textwidth}
      \centering
      \includegraphics[width=\linewidth]{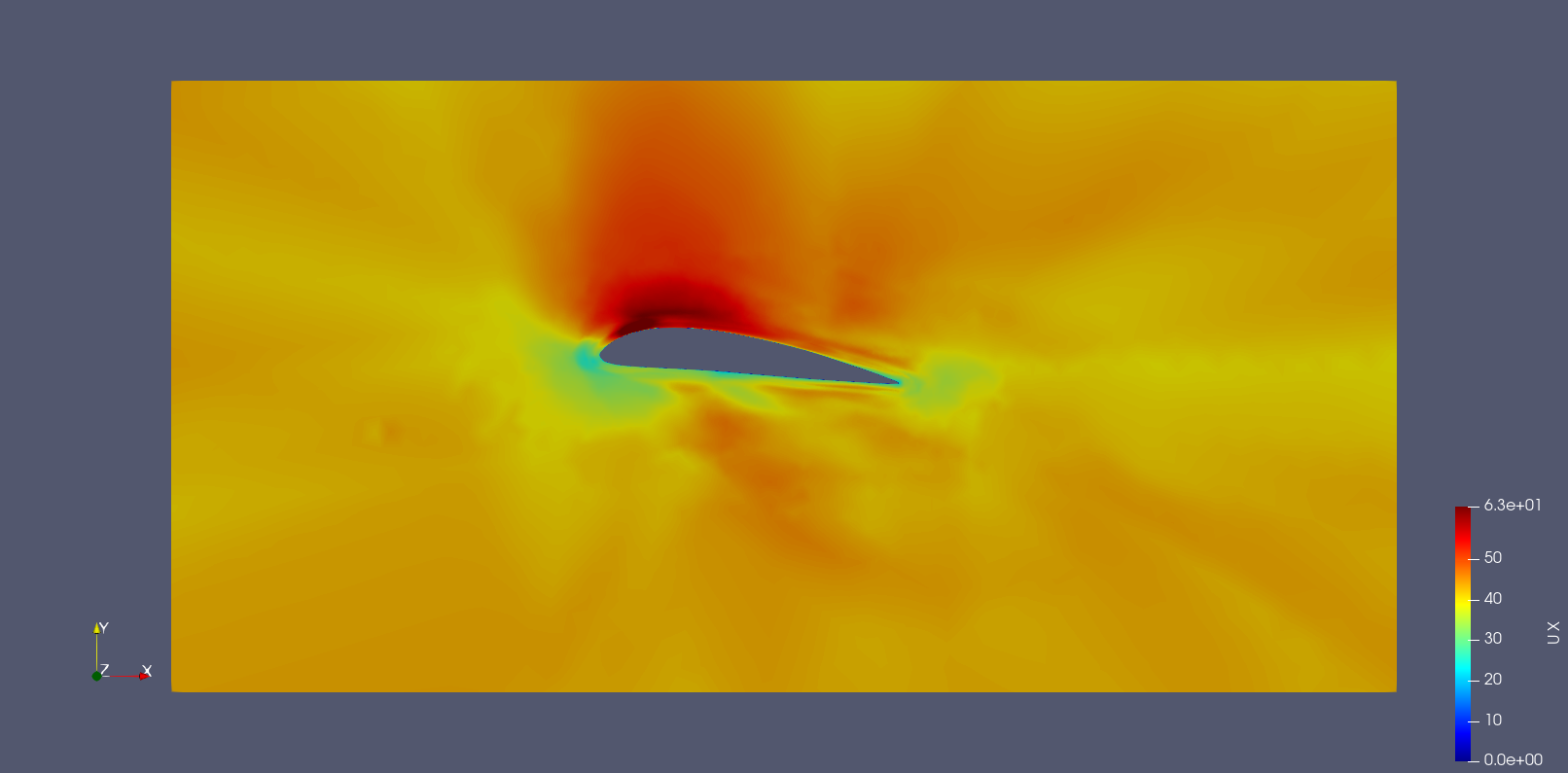}  
      \caption{Sampling + Radius graph}
      \label{fig:SAGE_radius}
    \end{subfigure}
    
    \caption{Comparison of the $x$-component of the velocity field for the two different GraphSAGE models.}
    \label{fig:SAGE}
\end{figure}

\begin{figure}
    \begin{subfigure}{\textwidth}
        \centering
        \includegraphics[width=\linewidth]{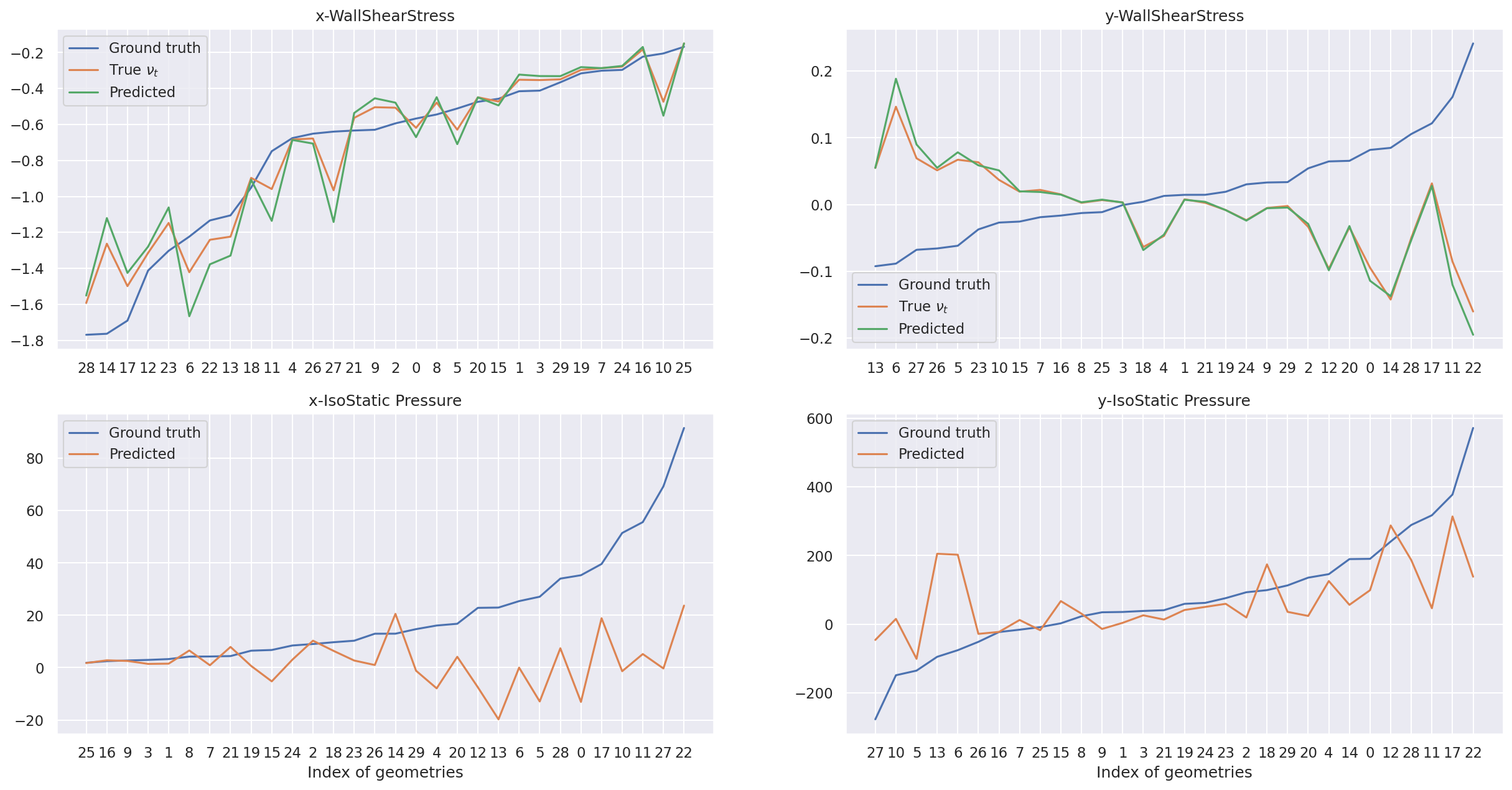}
         \caption{CFD mesh}
    \end{subfigure}
    
    \begin{subfigure}{\textwidth}
        \centering
        \includegraphics[width=\linewidth]{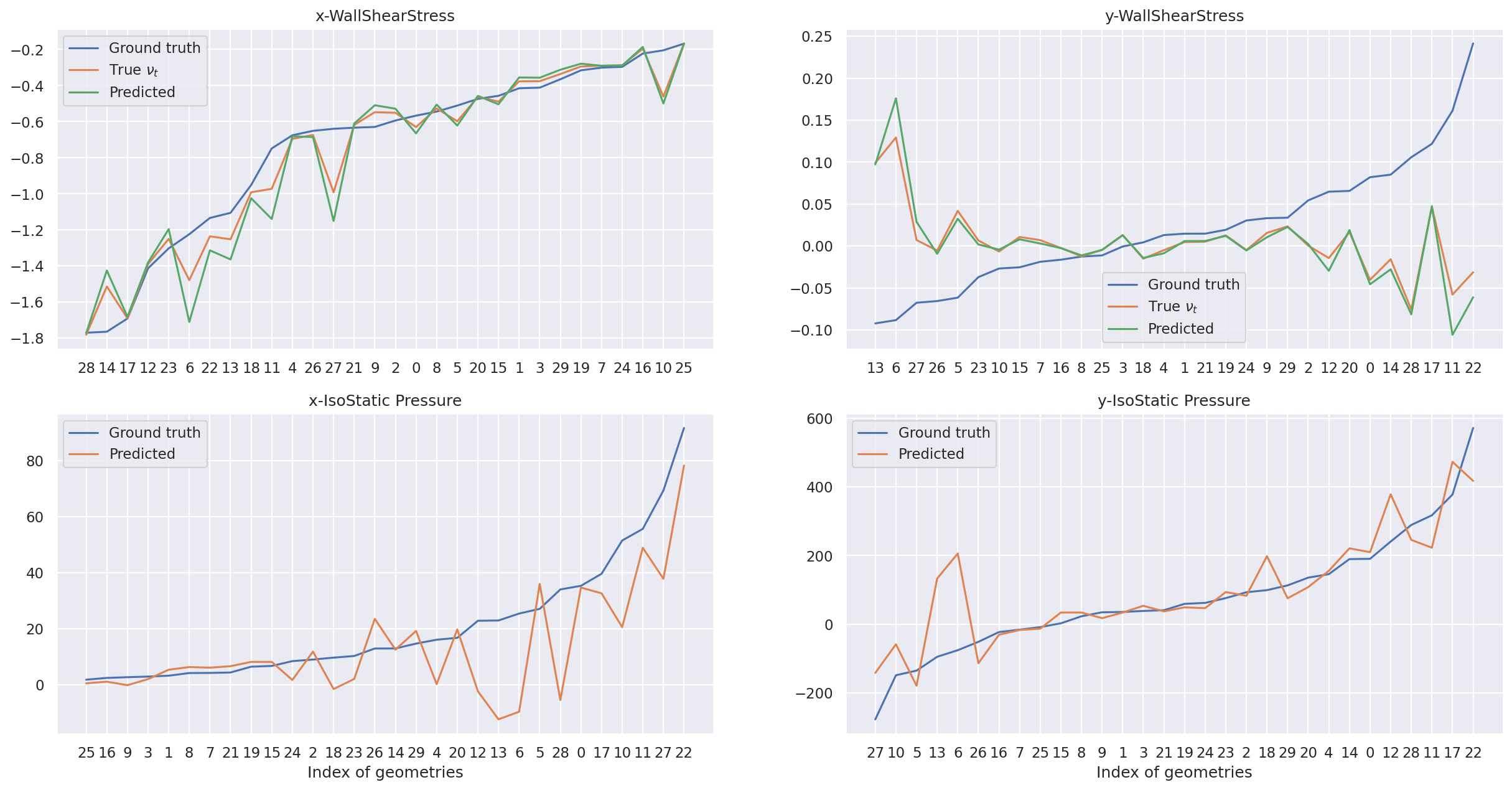}
         \caption{Sampling + Radius graph}
    \end{subfigure}
    
    \caption{Global coefficients over the test set for the GraphSAGE models.}
    \label{fig:SAGE_global}
\end{figure}

The $x$ component of the velocity field is shown in figure \ref{fig:SAGE} over a random test example that we will keep for the rest of the report. Ultimately, global coefficients over the test set are given in figure \ref{fig:SAGE_global}, in the plots for the Waal Shear Stress components, the green curve labelled "True $\nu_t$" is the results when we use the predicted Jacobian of the velocity field with the true $\nu_t$. It allows us to know if the error comes from the Jacobian or from the turbulent viscosity prediction. As expected, the model trained with a fixed radius outperforms the model trained over the CFD mesh.

\subsubsection*{GAT}
The GAT layer \cite{GAT} is for inductive tasks too but is often better performing than GraphSAGE. We used a 4-layer GAT network with 4 heads, 64 channels and ReLU activation. The optimization is done with Adam, a one-cycle cosine learning rate of maximum $10^{-3}$, a batch size of 1 and during 1000 epochs.

\begin{figure}
    \begin{subfigure}{\textwidth}
        \centering
        \includegraphics[width=\linewidth]{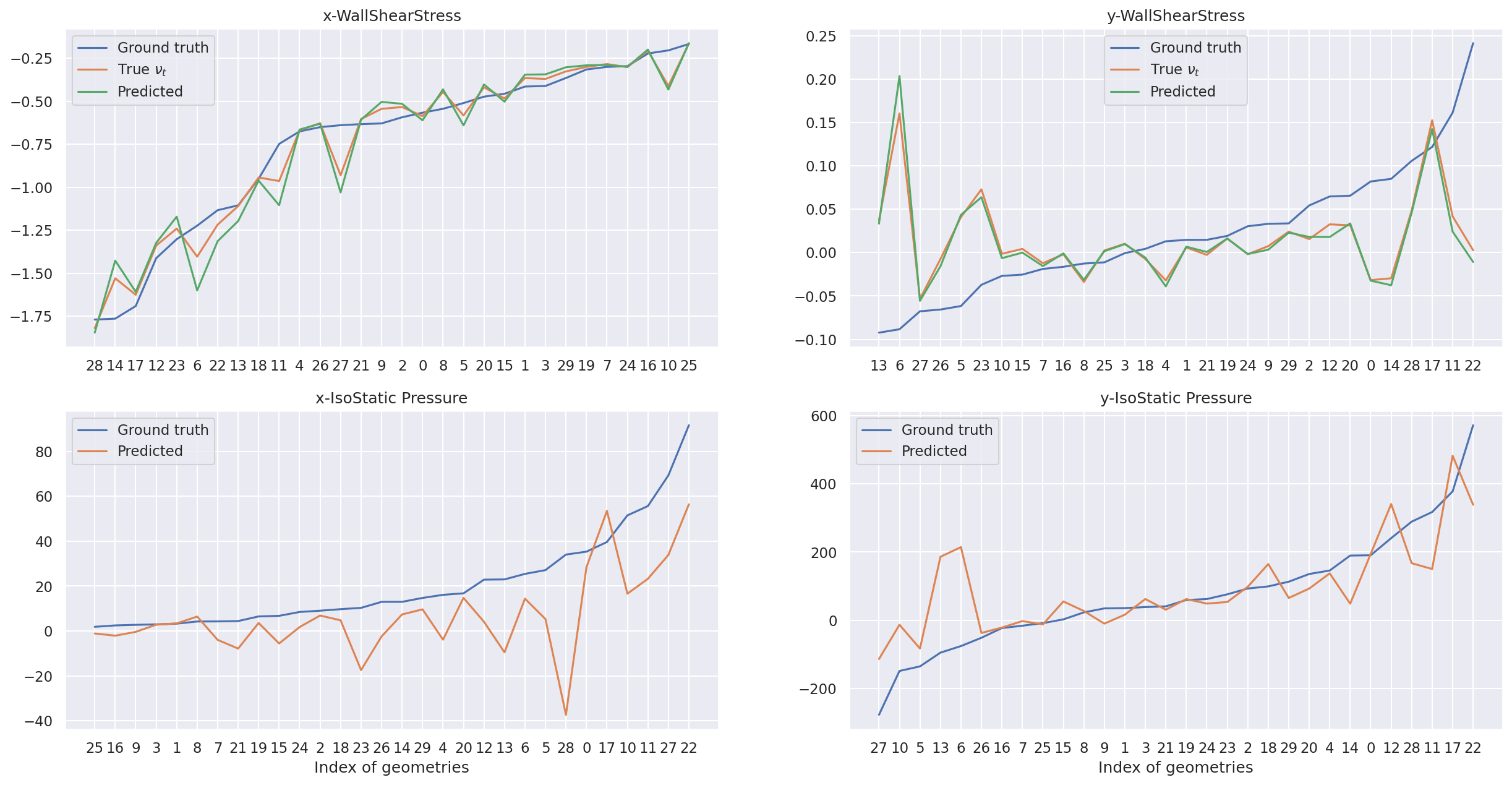}
         \caption{CFD mesh}
    \end{subfigure}
    
    \begin{subfigure}{\textwidth}
        \centering
        \includegraphics[width=\linewidth]{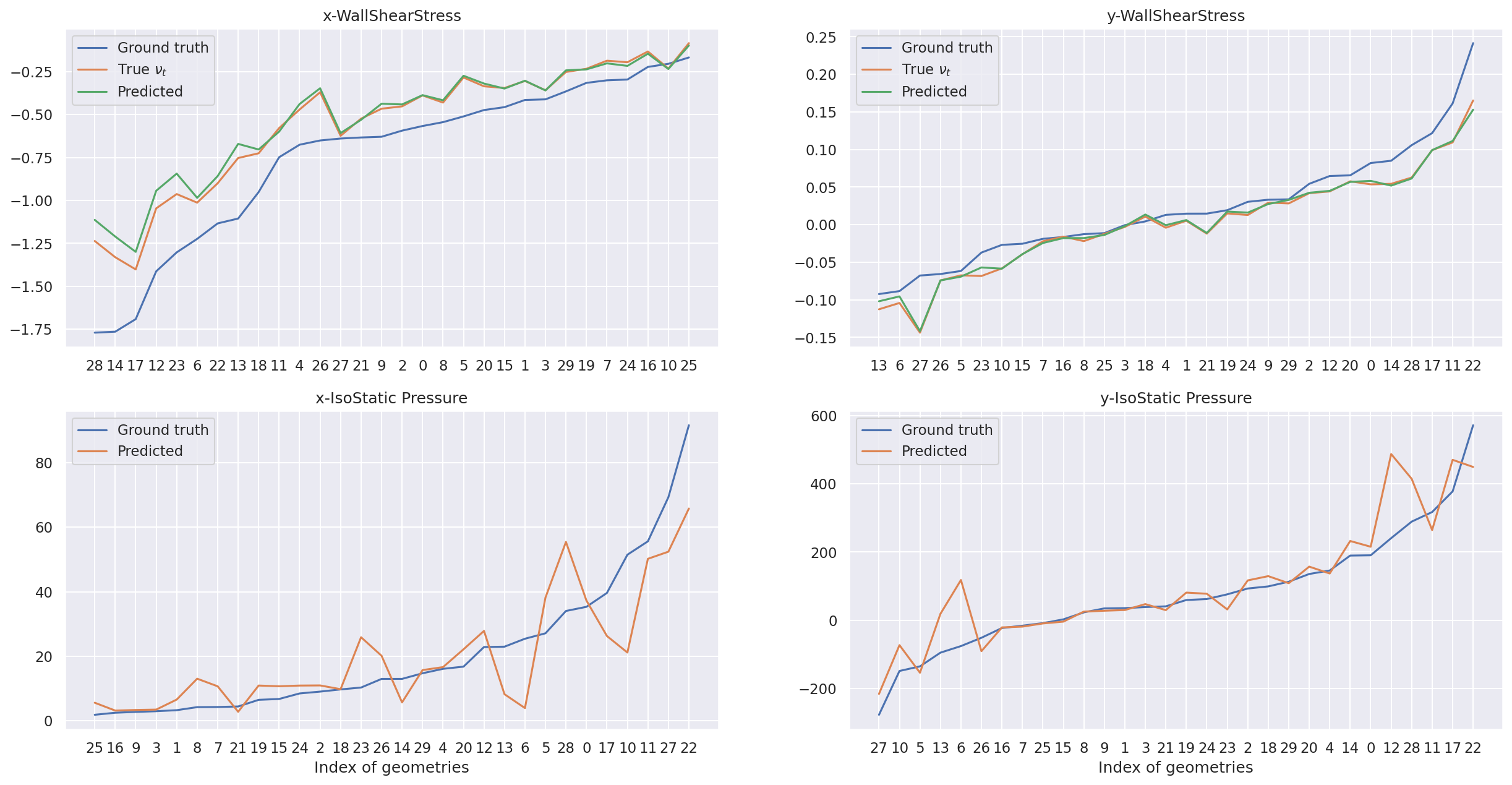}
         \caption{Sampling + Radius graph}
    \end{subfigure}
    
    \caption{Global coefficients over the test set for the GAT models.}
    \label{fig:GAT_global}
\end{figure}

The experiments are done in the same way as in the GraphSAGE section. Table \ref{GAT_score} gives the scores of the GAT mode. As for the GraphSAGE mode, the radius graph method outperforms the CFD mesh approach. Moreover, it is more robust to change of the number of points and to the rotation of the airfoil. As expected, the GAT model outperforms the GraphSAGE model. However, we remark on figure \ref{fig:GAT_global} that the model trained with the trained graph has a hard time to predict the velocity field at the surface of the geometry which is not the case with the GraphSAGE model or with the GAT model trained on the CFD mesh.

\begin{table}
    \centering
    \begin{tabular}{|c|c|c|c|c|c|c|c|}
        \hline
        \multicolumn{3}{|c|}{} & Val & Test & Test noise & Test rot & Test big \\
        \hline
        \multirow{6}{*}{CFD Mesh} & \multicolumn{2}{c|}{$\mathcal{L}_\mathcal{V}$} & 0.299 & 0.166 & 0.176 & 1.52 & 0.486 \\
         & \multicolumn{2}{c|}{$\mathcal{L}_\mathcal{S}$} & 0.424 & 0.225 & 0.261 & 0.926 & 0.622 \\
         \cline{2-8}
         & \multirow{4}{*}{glob. MAE} & x-WSS & 0.064 & 0.109 & 0.117 & 0.396 & 0.211 \\
         & & y-WSS & 0.095 & 0.059 & 0.055 & 0.196 & 0.160 \\
         & & x-IsoP & 25.3 & 15.1 & 15.9 & 93.5 & 49.4 \\
         & & y-IsoP & 72.9 & 72.8 & 76.7 & 160 & 125 \\
         \hline
         \multirow{6}{*}{Sampled radius graph} & \multicolumn{2}{c|}{$\mathcal{L}_\mathcal{V}$} & 0.123 & 0.080 & 0.091 & 0.897 & 0.193 \\
         & \multicolumn{2}{c|}{$\mathcal{L}_\mathcal{S}$} & 0.443 & 0.262 & 0.283 & 0.841 & 0.499 \\
         \cline{2-8}
         & \multirow{4}{*}{glob. MAE} & x-WSS & 0.228 & 0.214 & 0.193 & 0.150 & 0.485 \\
         & & y-WSS & 0.025 & 0.019 & 0.018 & 0.045 & 0.050 \\
         & & x-IsoP & 10.5 & 8.10 & 8.76 & 57.6 & 12.7 \\
         & & y-IsoP & 35.2 & 47.5 & 47.7 & 91.4 & 79.3 \\
         \hline
    \end{tabular}
    \caption{\label{GAT_score} Scores of the GAT model.}
\end{table}

\begin{figure}
    \begin{subfigure}{\textwidth}
      \centering
      \caption{Ground truth}
      \includegraphics[width=\linewidth]{Baselines/vx_ground_truth.png}  
    \end{subfigure}

    \begin{subfigure}{.5\textwidth}
      \centering
      \includegraphics[width=\linewidth]{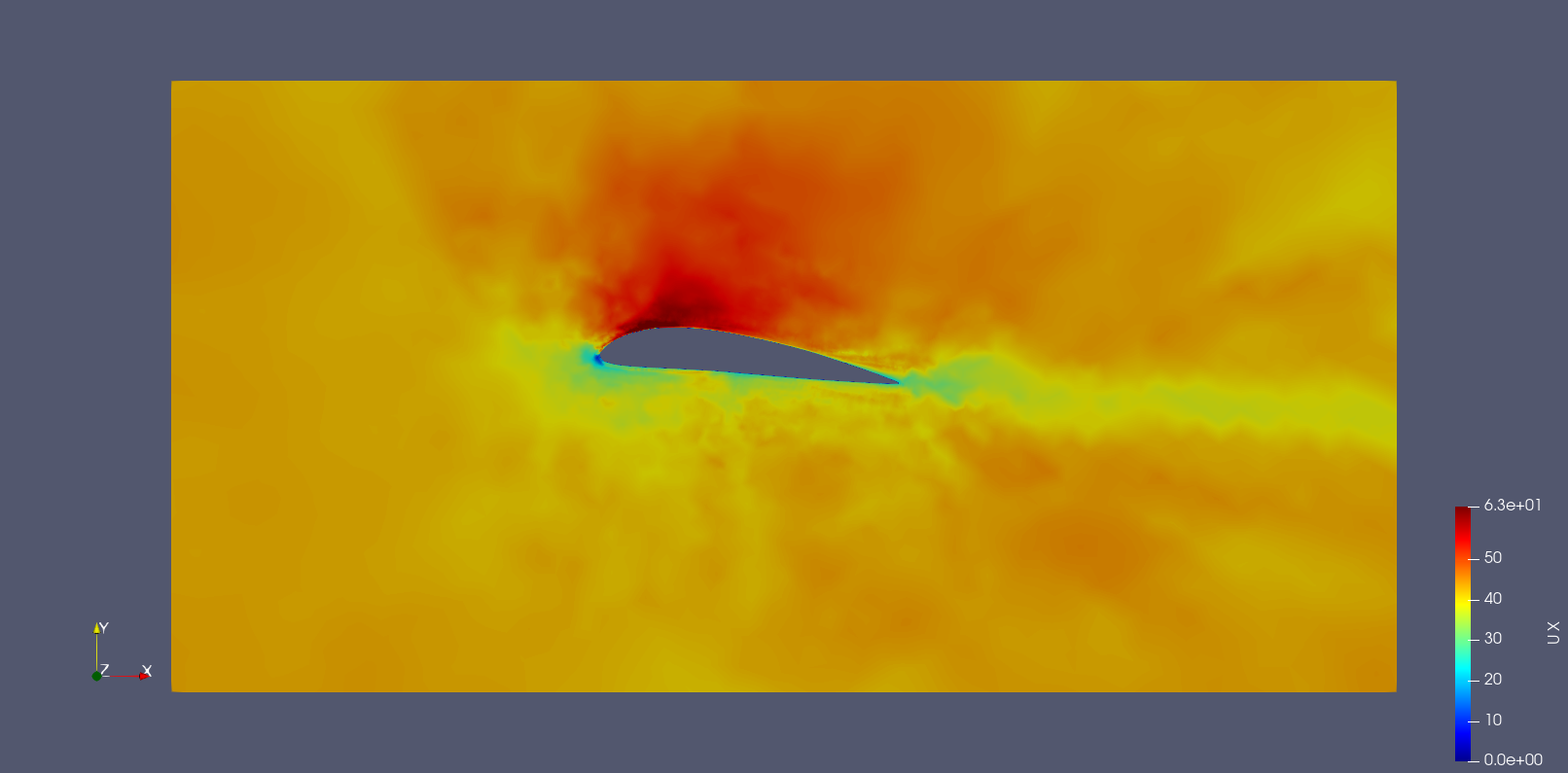}  
      \caption{CFD mesh}
    \end{subfigure}
    \begin{subfigure}{.5\textwidth}
      \centering
      \includegraphics[width=\linewidth]{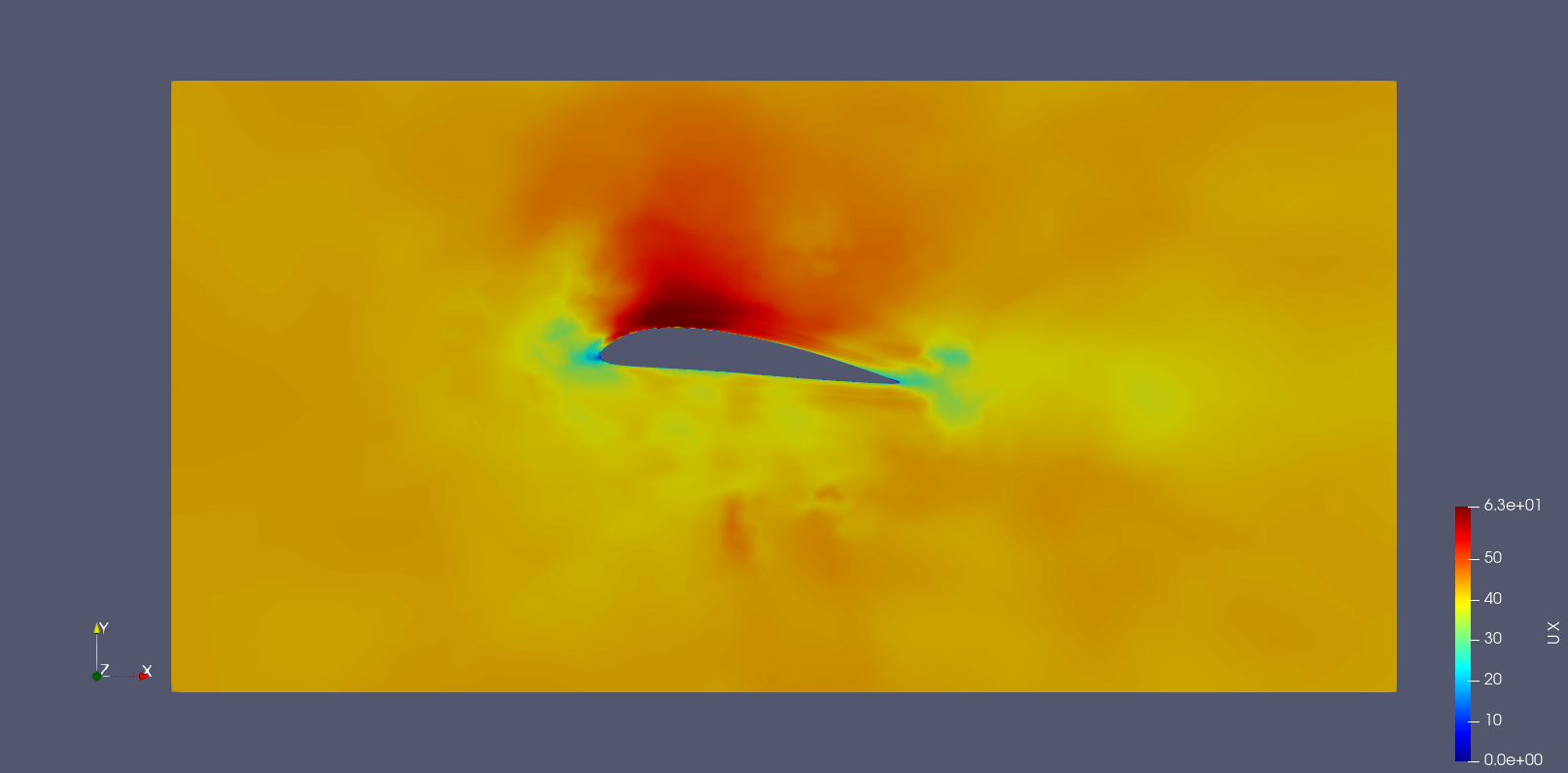}  
      \caption{Sampling + Radius graph}
    \end{subfigure}
    
    \caption{Comparison of the $x$-component of the velocity field for the two different GAT models.}
    \label{fig:GAT}
\end{figure}

\subsubsection*{PointNet}
PointNet \cite{pointnet} is not a GNN but a model that performs on point clouds. It is a two scales architecture that is better equipped to handle long-range interactions between nodes compared to previous GNN. Moreover, it can be built with a GraphSAGE or a GAT layer for the representation task. We chose to keep an MLP for the representation task as it gives similar scores and is faster to train. For the architecture, see \cite{pointnet} for the segmentation task, we just changed the output in order to fit with our problem. The optimization is done with Adam, a one-cycle cosine learning rate of maximum $10^{-3}$, a batch size of 1 and during 1000 epochs.

\begin{figure}
    \begin{subfigure}{\textwidth}
      \centering
      \caption{Ground truth}
      \includegraphics[width=\linewidth]{Baselines/vx_ground_truth.png}  
    \end{subfigure}

    \begin{subfigure}{.5\textwidth}
      \centering
      \includegraphics[width=\linewidth]{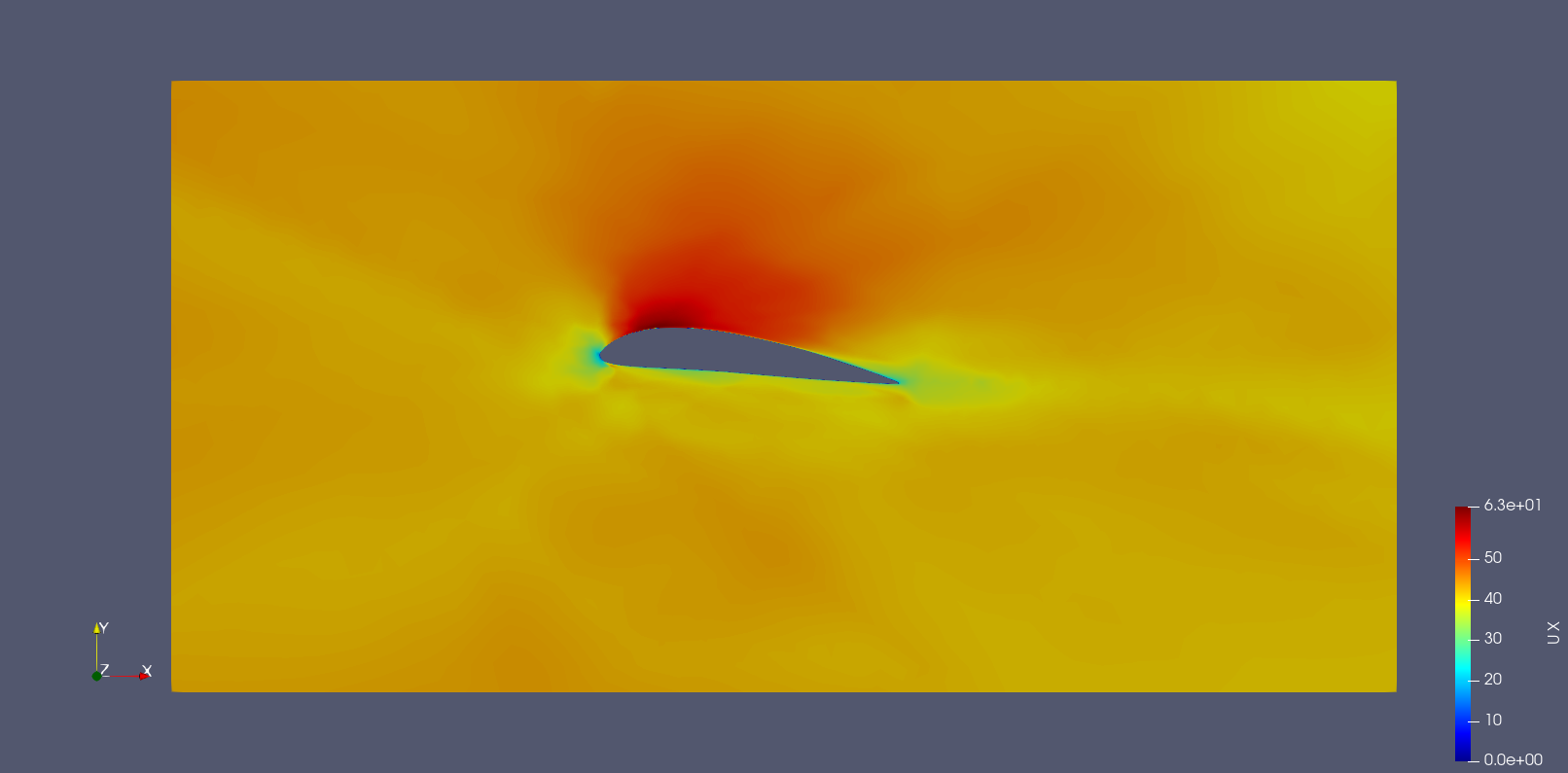}  
      \caption{CFD mesh}
    \end{subfigure}
    \begin{subfigure}{.5\textwidth}
      \centering
      \includegraphics[width=\linewidth]{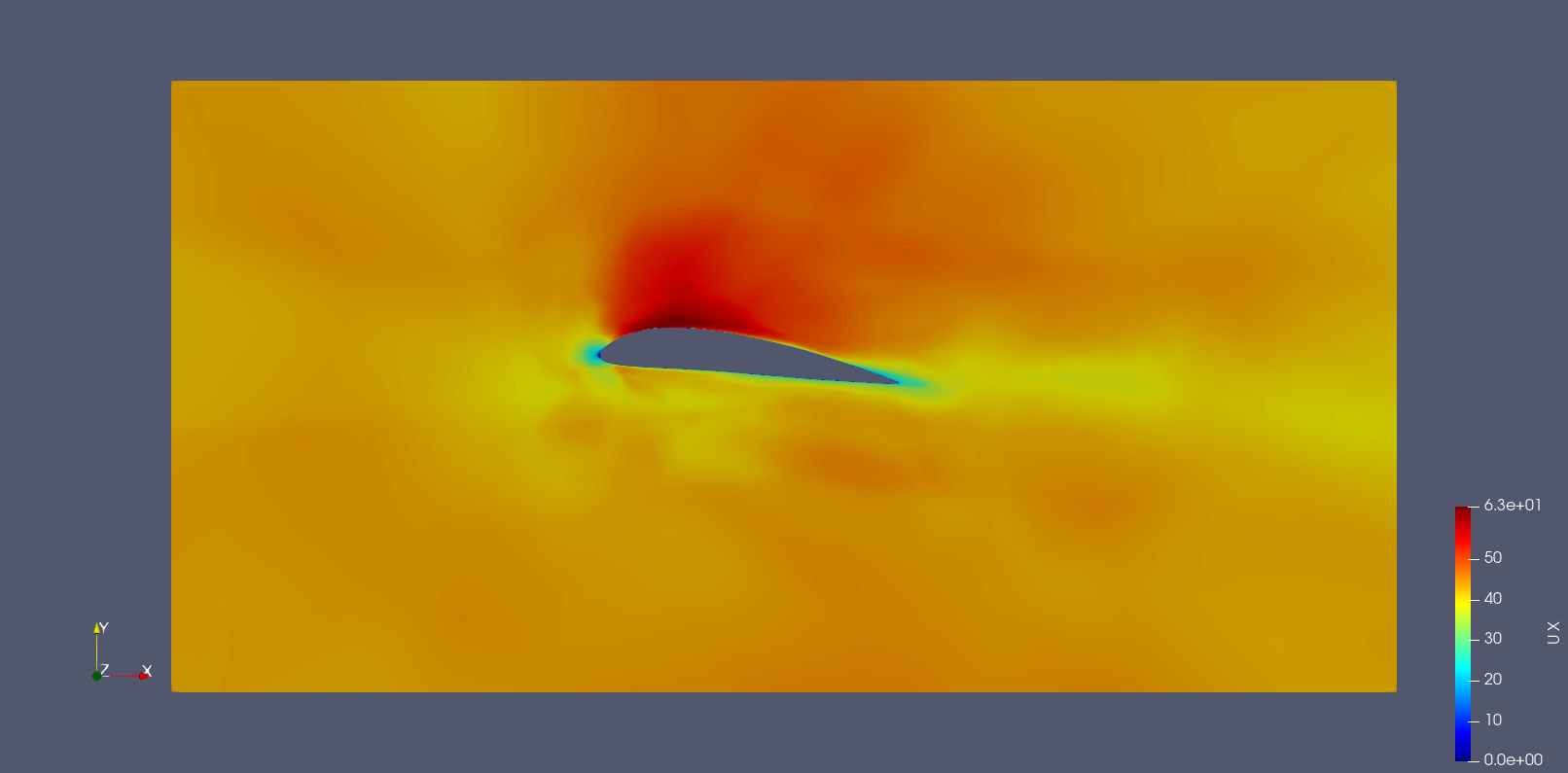}  
      \caption{Sampling + Radius graph}
    \end{subfigure}
    
    \caption{Comparison of the $x$-component of the velocity field for the two different PointNet models.}
    \label{fig:PN}
\end{figure}

\begin{figure}
    \begin{subfigure}{\textwidth}
        \centering
        \includegraphics[width=\linewidth]{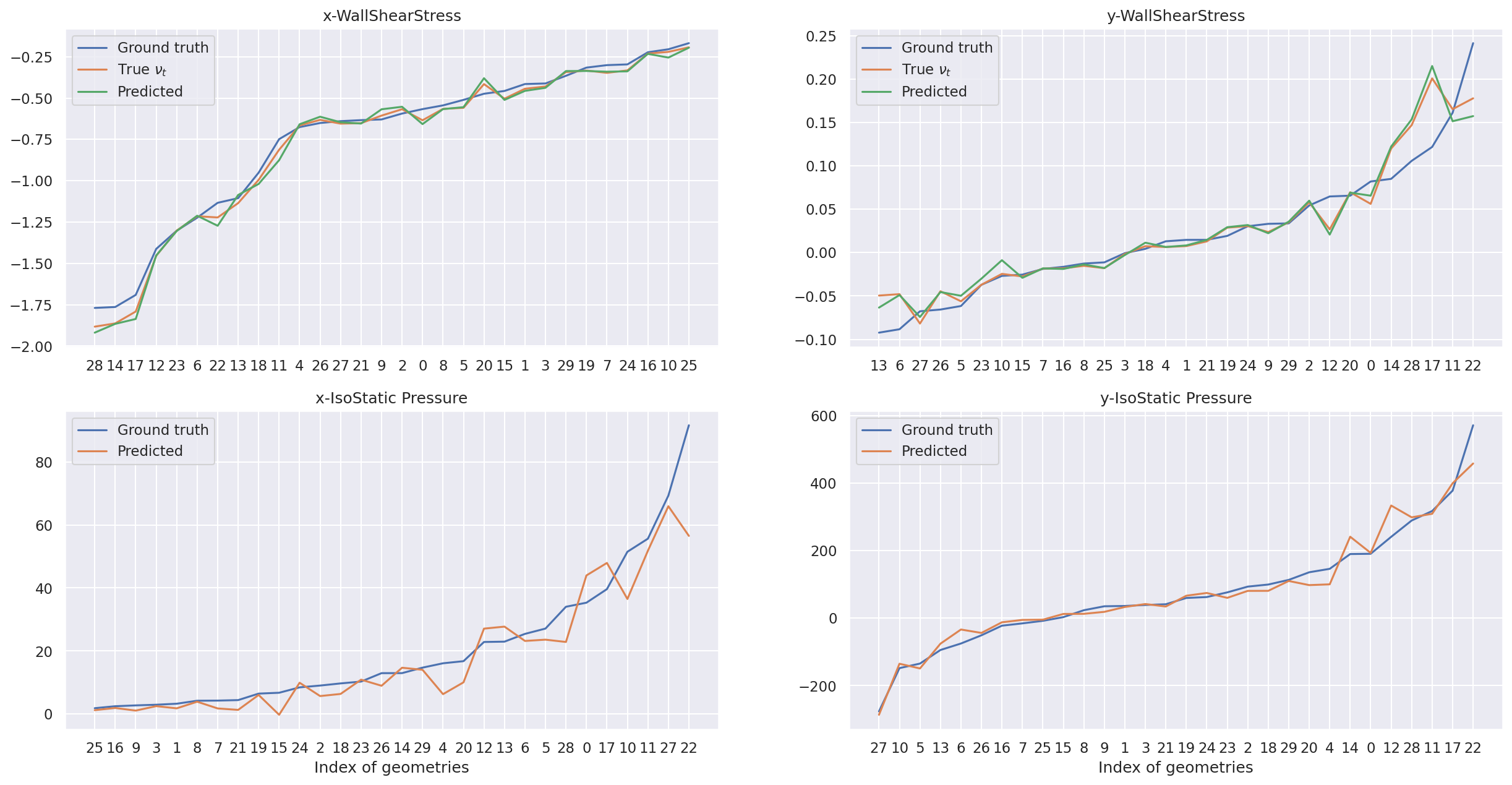}
         \caption{CFD mesh}
    \end{subfigure}
    
    \begin{subfigure}{\textwidth}
        \centering
        \includegraphics[width=\linewidth]{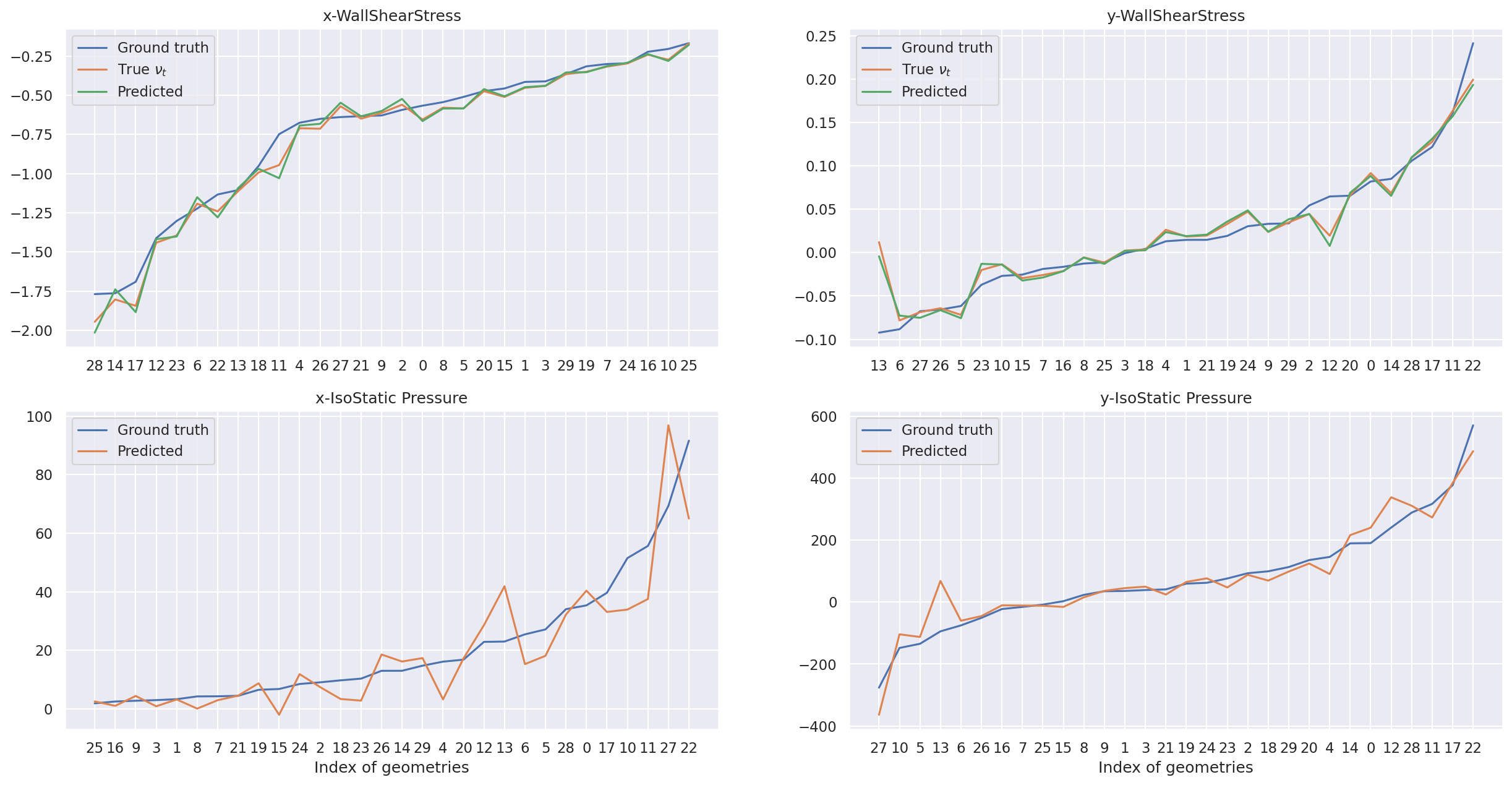}
         \caption{Sampling + Radius graph}
    \end{subfigure}
    
    \caption{Global coefficients over the test set for the PointNet models.}
    \label{fig:PN_global}
\end{figure}

Table \ref{PN_score} gives the scores for the PointNet model, as expected, it is the best of our baselines and the radius graph method gives similar results compared to the CFD mesh one. Figure \ref{fig:PN} shows the $x$-component of the velocity field for the different PointNet models and figure \ref{fig:PN_global} the global coefficients associated with the test set samples.


\begin{table}
    \centering
    \begin{tabular}{|c|c|c|c|c|c|c|c|}
        \hline
        \multicolumn{3}{|c|}{} & Val & Test & Test noise & Test rot & Test big \\
        \hline
        \multirow{6}{*}{CFD Mesh} & \multicolumn{2}{c|}{$\mathcal{L}_\mathcal{V}$} & 0.061 & 0.032 & 0.044 & 1.30 & 0.077 \\
         & \multicolumn{2}{c|}{$\mathcal{L}_\mathcal{S}$} & 0.138 & 0.069 & 0.094 & 2.75 & 0.112 \\
         \cline{2-8}
         & \multirow{4}{*}{glob. MAE} & x-WSS & 0.048 & 0.053 & 0.081 & 0.393 & 0.130 \\
         & & y-WSS & 0.021 & 0.018 & 0.018 & 0.098 & 0.050 \\
         & & x-IsoP & 10.7 & 4.99 & 5.06 & 117 & 11.9 \\
         & & y-IsoP & 30.4 & 21.0 & 23.5 & 87.0 & 39.3 \\
         \hline
         \multirow{6}{*}{Sampled radius graph} & \multicolumn{2}{c|}{$\mathcal{L}_\mathcal{V}$} & 0.075 & 0.036 & 0.047 & 1.11 & 0.128 \\
         & \multicolumn{2}{c|}{$\mathcal{L}_\mathcal{S}$} & 0.117 & 0.118 & 0.108 & 0.873 & 0.175 \\
         \cline{2-8}
         & \multirow{4}{*}{glob. MAE} & x-WSS & 0.048 & 0.062 & 0.091 & 0.338 & 0.160 \\
         & & y-WSS & 0.016 & 0.014 & 0.016 & 0.052 & 0.050 \\
         & & x-IsoP & 10.4 & 7.11 & 6.97 & 65.2 & 17.8 \\
         & & y-IsoP & 35.8 & 30.6 & 32.8 & 80.1 & 69.9 \\
         \hline
    \end{tabular}
    \caption{\label{PN_score} Scores of the PointNet model.}
\end{table}

To summarize, table \ref{tab:Summary_baselines} gives the total MSE $\mathcal{L}$ between the one trained with the radius graph method and their number of parameters plus their training time.

\begin{table}
    \centering
    \begin{tabular}{|c|c|c|c|c|c|c|c|}
         \hline
        & Val & Test & Test noise & Test rot & Test big & \#Parameters & Training time \\
         \hline
        GraphSAGE & 0.751 & 0.403 & 0.436 & 3.09 & 1.59 & 17604 & 0:43.47 \\
        \hline
        GAT & 0.566 & 0.342 & 0.437 & 2.45 & 1.11 & 138532 & 1:14.38 \\
        \hline
        PointNet & 0.192 & 0.154 & 0.155 & 1.98 & 0.303 & 64892 & 0:47.11 \\
        \hline
    \end{tabular}
    \caption{Summary of the MSE for the different baselines.}
    \label{tab:Summary_baselines}
\end{table}

\subsection{Architecture proposed}
\emph{The architecture proposed in this section is an adaptation of the Graph Kernel Network (GKN) presented in the Neural Operator papers \cite{GKN, MGKN}}

We extend the inputs to a 6-dimensional vector by adding two components set to 0 in place of the pressure and $y$-component velocity, this for coding ease. We do not use the CFD mesh and we follow the same procedure as for the baselines by sampling 1600 points and building a radius graph of radius 0.1. 

\subsubsection{GKN ReLU}
We encode this 6-dimensional inputs by a 3-layers neural network $\phi_{\theta} : \mathbb{R}^6 \to \mathbb{R}^r$, where $r$ is the dimension of the representation (the network has $6-64-64-r$ neurons and a ReLU activation). Then we use the graph to construct a convolution out of it with a kernel $\kappa_\Theta : \mathbb{R}^8 \to \mathcal{M}_r$ that takes in input the attribute of the edge of the graph and outputs a squared matrix of the size of the representation. The attributes of an edge $e_{ij}$ is the relative position, the relative velocity and the relative pressure between the nodes $i$ and $j$ plus the signed distance function at nodes $i$ and $j$ and the velocity at the inlet (which gives an 8-dimensional vector). This convolution can be composed with an activation function $\sigma$ and a residue and iterated. After those convolutions, we decode the signal via a 3-layers neural network $\psi_\gamma : \mathbb{R}^r \to \mathbb{R}^4$ (the network has $r-64-64-4$ neurons and a ReLU activation) to find the predicted signal. If $h_i\in\mathbb{R}^4$ is our predicted signal at node $i$, $x_i\in \mathbb{R}^6$ our input and $T$ the number of iterations, we have:
\begin{align*}
    \begin{cases}
        h_i^0 = \phi_\theta(x_i) \\
        h_i^t = \sigma\left( \frac{1}{|\mathcal{N}_i|}\sum_{j\in\mathcal{N}_i} \kappa_\Theta(e^t_{ij}) h^{t-1}_j  \right) + h_i^{t-1}, \qquad t = 1,\,\dots,\, T-1 \\
        h_i = h_i^T = \psi_\gamma \left( \frac{1}{|\mathcal{N}_i|}\sum_{j\in\mathcal{N}_i} \kappa_\Theta(e^{T-1}_{ij}) h_j^{T-1} + h_i^{T-1} \right)
    \end{cases}
\end{align*}
where $\mathcal{N}_i$ is the set of neighbours of the node $i$ (which includes $i$ itself). We also remark that the edge attribute $e^t_{ij}$ depends on the iteration $t$, this is because we update the edge attribute at each iteration, meaning for two index $i$ and $j$ and $x$ the input: 
\begin{align*}
    e^0_{ij} &= \left( (x_i - x_j)_1,\, (x_i - x_j)_2,\, (x_i - x_j)_3,\, (x_i - x_j)_4,\, (x_i - x_j)_5,\, (x_i)_6,\, (x_j)_6,\, v_{inlet}\right) \\
    e^t_{ij} &= \left( (x_i - x_j)_1,\, (x_i - x_j)_2,\, (\psi_\gamma(h^t_i) - \psi_\gamma(h^t_j))_1,\, (\psi_\gamma(h^t_i) - \psi_\gamma(h^t_j))_2,\, (\psi_\gamma(h^t_i) - \psi_\gamma(h^t_j))_3,\, g_i,\, g_j,\, v_{inlet}\right)
\end{align*}
for $t = 1,\,\dots,\, T-1$. Where $v_{inlet}$ is the velocity at the inlet of the domain and where the components of an input $x_i$ at node $i$ are defined as:
\begin{itemize}
    \item $(x_i)_1$ : $x$-component of the position of the node $i$,
    \item $(x_i)_2$ : $y$-component of the position of the node $i$,
    \item $(x_i)_3$ : $x$-component of the flow velocity at node $i$,
    \item $(x_i)_4$ : $y$-component of the flow velocity at node $i$,
    \item $(x_i)_5$ : flow pressure at node $i$,
    \item $(x_i)_6$ : distance from the airfoil of the node $i$.
\end{itemize}

Let us now explain why such architecture. The idea behind it is to first, try to linearize the PDE by finding an appropriate representation of the inputs as it is suggested in the example of the Hopf-Cole transformation or in the Koopman theory \cite{koopman} we talked about in section \ref{PDE} and to inverse this representation at the end of the network (through $\phi_\theta$ and $\psi_\gamma$). After representing it, we would like to treat it as a linear elliptic PDE, and find its solution by doing a convolution over the represented inputs. This convolution at node $i$ is approximated via a Monte-Carlo estimation by averaging over the set of neighbours of the node $i$. This approximation underlined the assumption that the meaningful interactions between nodes are at most of radius 0.1. The kernel of this convolution is given by $\kappa_\Theta$. However, we should have put the representation of node $i$ or $j$ as input of the kernel to be consistent but we did not do that for two reasons. First, it allows us to hard-code some symmetries of the problem in the kernel, for example, the kernel should not depends on the absolute positions but on the relative positions. We can say the same thing for the velocity and the pressure even though the non linearity in the velocity variables in the RANS equations may suggest that the absolute velocity could play a role. This is why we kept the velocity at the inlet as input of the kernel. The signed distance function allows the kernel to have access to the information of the geometry. Second, this is faster in the training process not to use the representation as inputs of the kernel. Finally, the residue is here to help the optimization process.

A remark can be done on the Monte Carlo approximation of the convolution. For each training sample, the subsampling is uniform over the point cloud but is not uniform over the domain. There is a higher density of points close to the airfoil than far away from it. This means that the kernel $\kappa_\Theta$ is actually approximating a kernel divided by the probability density function of the point cloud distribution and that it should be sensitive to a change of distribution of the point cloud. We tried to make it more robust by applying a kernel density estimation to the point cloud and divide the kernel by it. This leads to unstable learning and numerical explosion. Hence, we kept the simple form of the convolution which is still adapted to the task we are looking at as all the test sets are sampled in the same manner. \\

Now, we present the results of this architecture. We set the size of the representation $r$ to 8 and the number of iterations $T$ to 5. Moreover, the kernel $\kappa_\Theta$ is defined as a 4-layer fully connected neural network with ReLU activation and 8-64-64-64-64 neurons (the output is then reshaped in a $8\times 8$ matrix). The optimization is done with Adam, a one-cycle cosine learning rate of maximum $10^{-3}$, a batch size of 1 and during 1000 epochs. 

\begin{figure}
    \begin{subfigure}{\textwidth}
      \centering
      \includegraphics[width=\linewidth]{Baselines/vx_ground_truth.png}
      \caption{Ground truth}
    \end{subfigure}

    \begin{subfigure}{\textwidth}
      \centering
      \includegraphics[width=\linewidth]{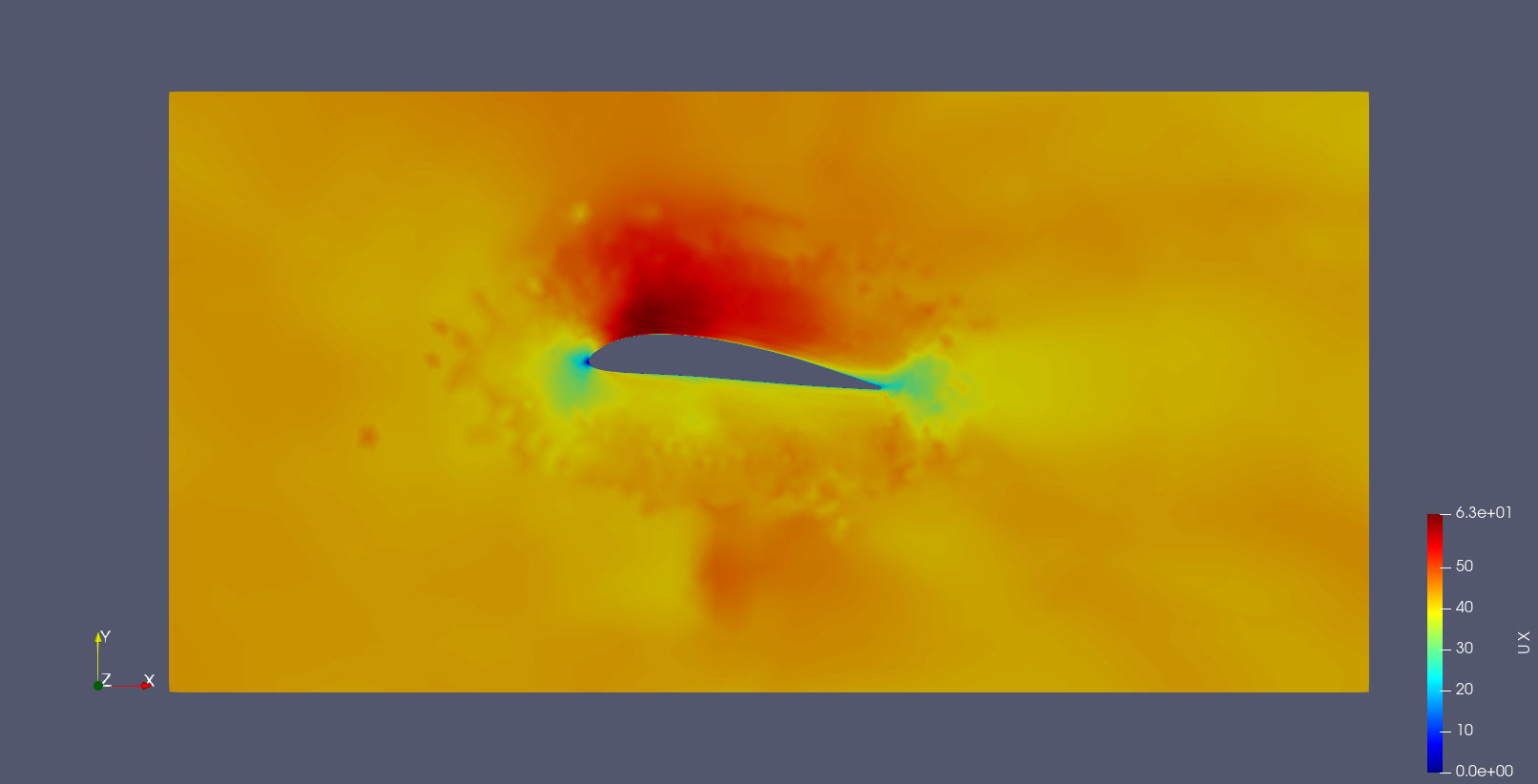}  
      \caption{Predicted}
    \end{subfigure}
    
    \caption{Comparison of the $x$-component of the velocity field for the GKN with ReLU kernel.}
    \label{fig:GKN}
\end{figure}

\begin{figure}
    \centering
    \includegraphics[width=\linewidth]{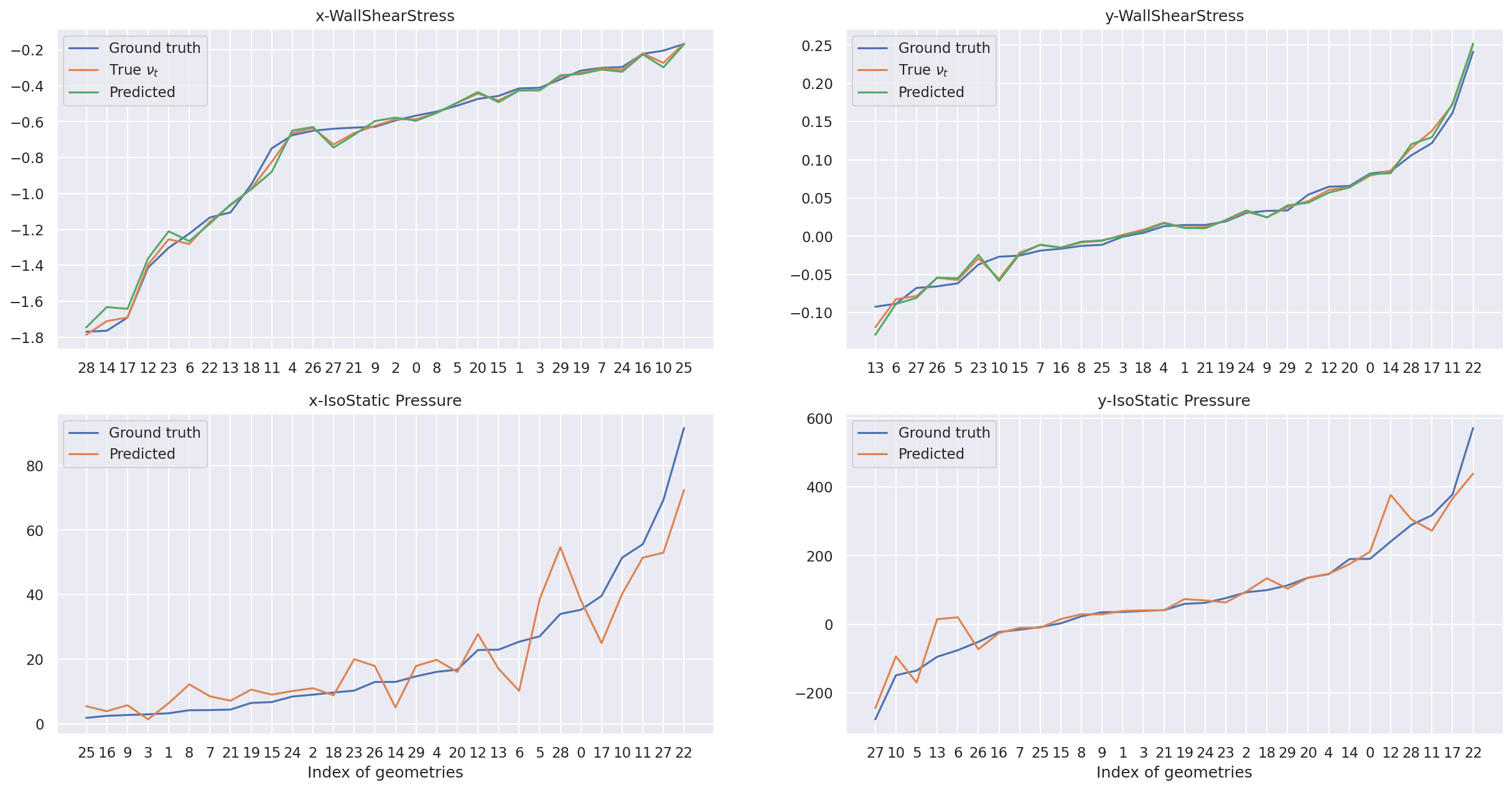}
    \caption{Global coefficients over the test set for the GKN with ReLU kernel.}
    \label{fig:GKN_global}
\end{figure}

Table \ref{GKN_score} gives the scores of the GKN with ReLU kernel. Figure \ref{fig:GKN} shows the $x$-component of the velocity field of the same test example as for the baselines and figure \ref{fig:GKN_global} gives the global coefficients curves over the test set for the GKN with ReLU kernel. We remark that the velocity field is not smooth and we can suspect the kernel not to have enough sensitivity with respect to its inputs. Actually, we can see $\kappa_\Theta$ as an implicit network encoding a "hyper-image" where a pixel is not a RGB channel but a $8\times 8$ matrix (or a 64-dimensional vector). New methods to better encode such signals have appeared recently, one of them is the SIREN \cite{siren} network. This is the topic of the next subsection.

\begin{table}
    \centering
    \begin{tabular}{|c|c|c|c|c|c|c|c|}
        \hline
        \multicolumn{2}{|c|}{} & Val & Test & Test noise & Test rot & Test big \\
        \hline
         \multicolumn{2}{|c|}{$\mathcal{L}_\mathcal{V}$} & 0.048 & 0.039 & 0.048 & 0.582 & 0.079 \\
         \multicolumn{2}{|c|}{$\mathcal{L}_\mathcal{S}$} & 0.077 & 0.048 & 0.071 & 0.308 & 0.091 \\
         \cline{1-7}
         \multirow{4}{*}{glob. MAE} & x-WSS & 0.020 & 0.040 & 0.054 & 0.109 & 0.086 \\
         & y-WSS & 0.008 & 0.008 & 0.010 & 0.053 & 0.023 \\
         & x-IsoP & 9.92 & 6.52 & 6.75 & 23.6 & 7.21 \\
         & y-IsoP & 29.7 & 28.2 & 29.8 & 45.4 & 45.1 \\
         \hline
    \end{tabular}
    \caption{\label{GKN_score} Scores of the GKN model.}
\end{table}

\subsubsection{GKN SIREN}
SIREN networks \cite{siren} are types of network that use a sine function as activation function and a well chosen initialization of their weights. They are built to overfit a signal in order to store an implicit representation of it. Common applications are the implicit representation of images or 3D shapes. They have shown to outperform greatly classical ReLU networks in this task by better representing high frequencies of the signal leading to a more detailed representation. Moreover, the derivatives of a SIREN network are SIREN networks and it was shown experimentally that they also accurately approximate the derivatives of the signal represented whereas ReLU network have constant by parts first derivatives and null higher order derivatives. This also allows SIREN networks to be trained through a loss over the derivatives of the networks. 

For us, SIREN networks seem appealing for two reasons. First, their ability to better represent signal might allow us to increase the scores of our GKN by replacing the ReLU network kernel by a SIREN kernel. Second, we might be able, with the help of the RANS equation, to find a PDE for the kernel itself and we could supervised directly the derivatives of the kernel. This second reason is definitely not straightforward and will not be discussed in the following even though it could be something to look at in a next work.

However, there is still a major problem with SIREN networks, they are built to overfit a signal and this is not suited for our task. We would like our network to be able to learn over different samples and to generalize. This can not be done using a SIREN network for the kernel, an extension of such technology to bypass this problem will be discuss in the next subsection.

We still test those two candidates (ReLU and SIREN kernels) in the overfitting setting. We choose a random sample of the training set and we trained both networks on it. The ReLU kernel is the same as we used previously and the SIREN kernel is a 4-layers kernel with the same amount of neurons as the ReLU one. What we remark is that the SIREN kernel does not need several iteration in the convolution process to give good results and we kept only one iteration in this setting. The optimization is done with Adam, a one-cycle cosine learning rate of maximum $10^{-3}$ and during 2000 epochs. Here the loss is the total MSE without separating volume and surface points.

\begin{table}
    \centering
    \begin{tabular}{|c|c|c|c|}
        \hline
        \multicolumn{2}{|c|}{} & ReLU & SIREN  \\
        \hline
         \multicolumn{2}{|c|}{$\mathcal{L}$} & $(1.51 \pm 0.39) \cdot 10^{-3}$ & $(7.14 \pm 2.22)\cdot 10^{-4}$ \\
         \cline{1-4}
         \multirow{4}{*}{glob. MAE} & x-WSS & $(7.34 \pm 3.47)\cdot 10^{-3}$ & $(7.24 \pm 4.66)\cdot 10^{-3}$  \\
         & y-WSS & $(4.54 \pm 3.15)\cdot 10^{-3}$ & $(1.35 \pm 0.97)\cdot 10^{-3}$ \\
         & x-IsoP & $(5.20 \pm 3.19)\cdot 10^{-1}$ & $(7.07 \pm 3.27)\cdot 10^{-1}$ \\
         & y-IsoP & $2.65 \pm 1.60$ & $2.35 \pm 2.03$ \\
         \hline
    \end{tabular}
    \caption{Scores of the GKN model with ReLU and SIREN kernel.}
    \label{SIREN_score}
\end{table}

\begin{figure}
    \begin{subfigure}{\textwidth}
      \centering
      \caption{Ground truth}
      \includegraphics[width=\linewidth]{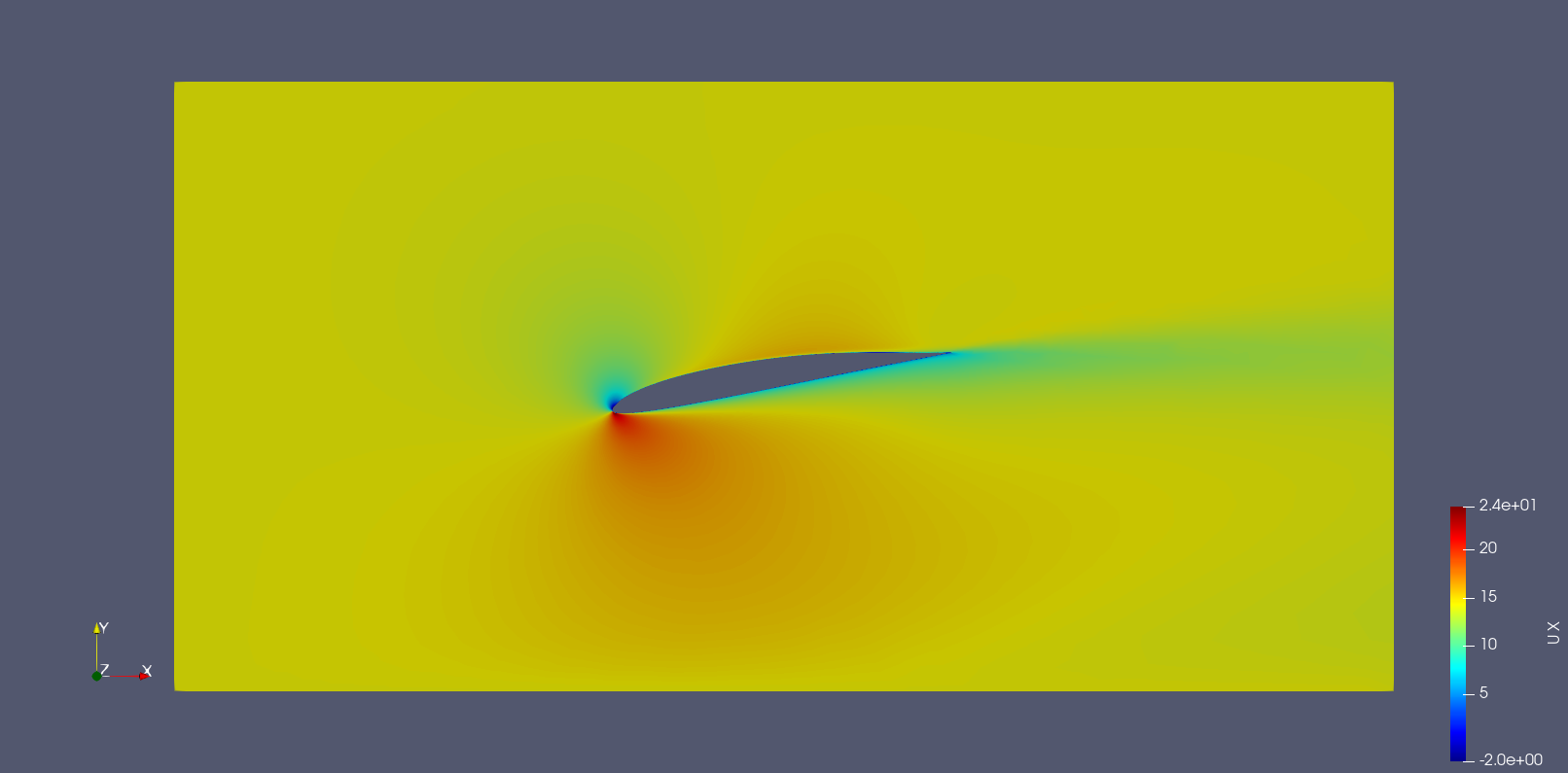}  
    \end{subfigure}

    \begin{subfigure}{.5\textwidth}
      \centering
      \includegraphics[width=\linewidth]{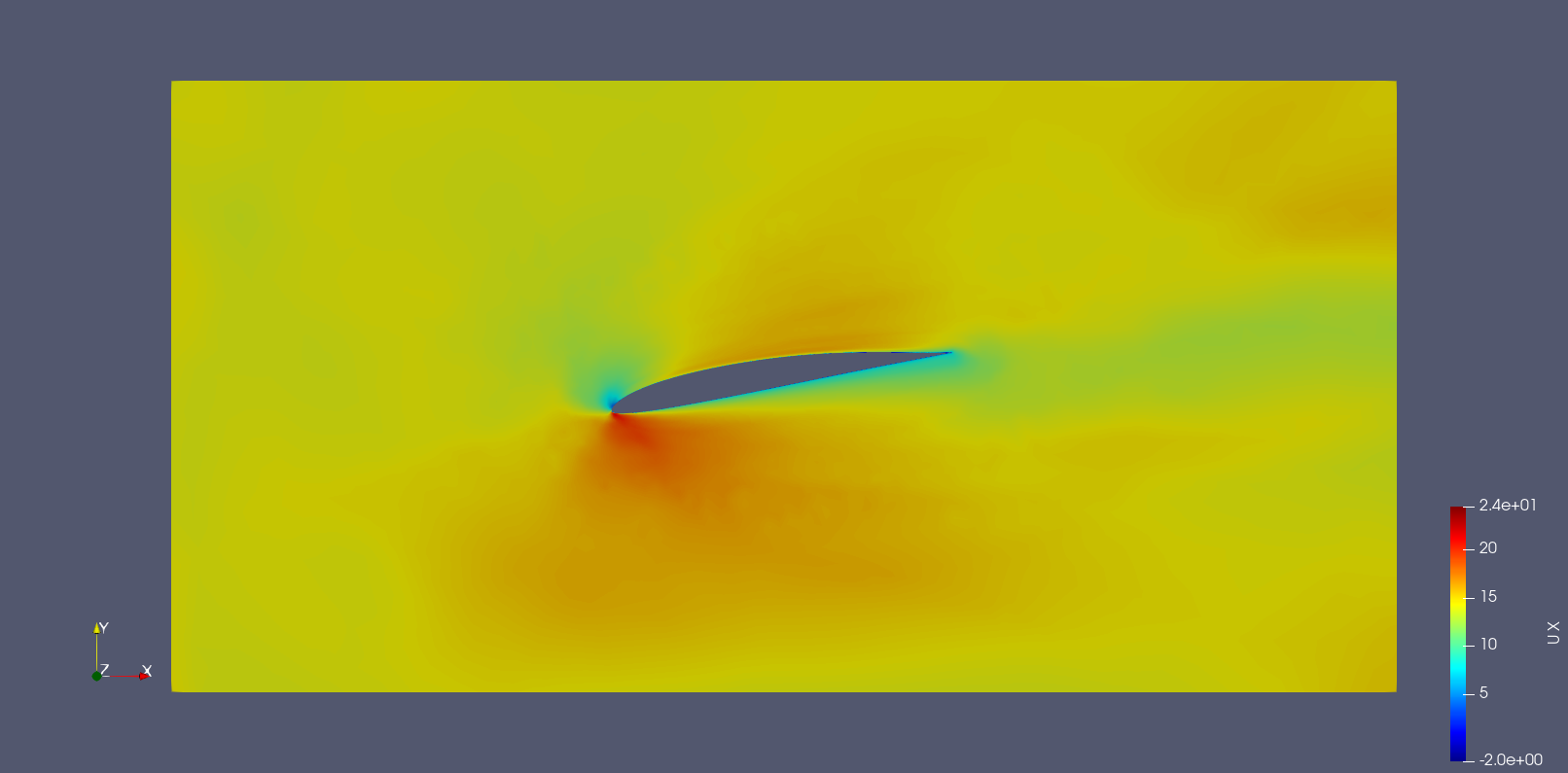}  
      \caption{ReLU kernel}
    \end{subfigure}
    \begin{subfigure}{.5\textwidth}
      \centering
      \includegraphics[width=\linewidth]{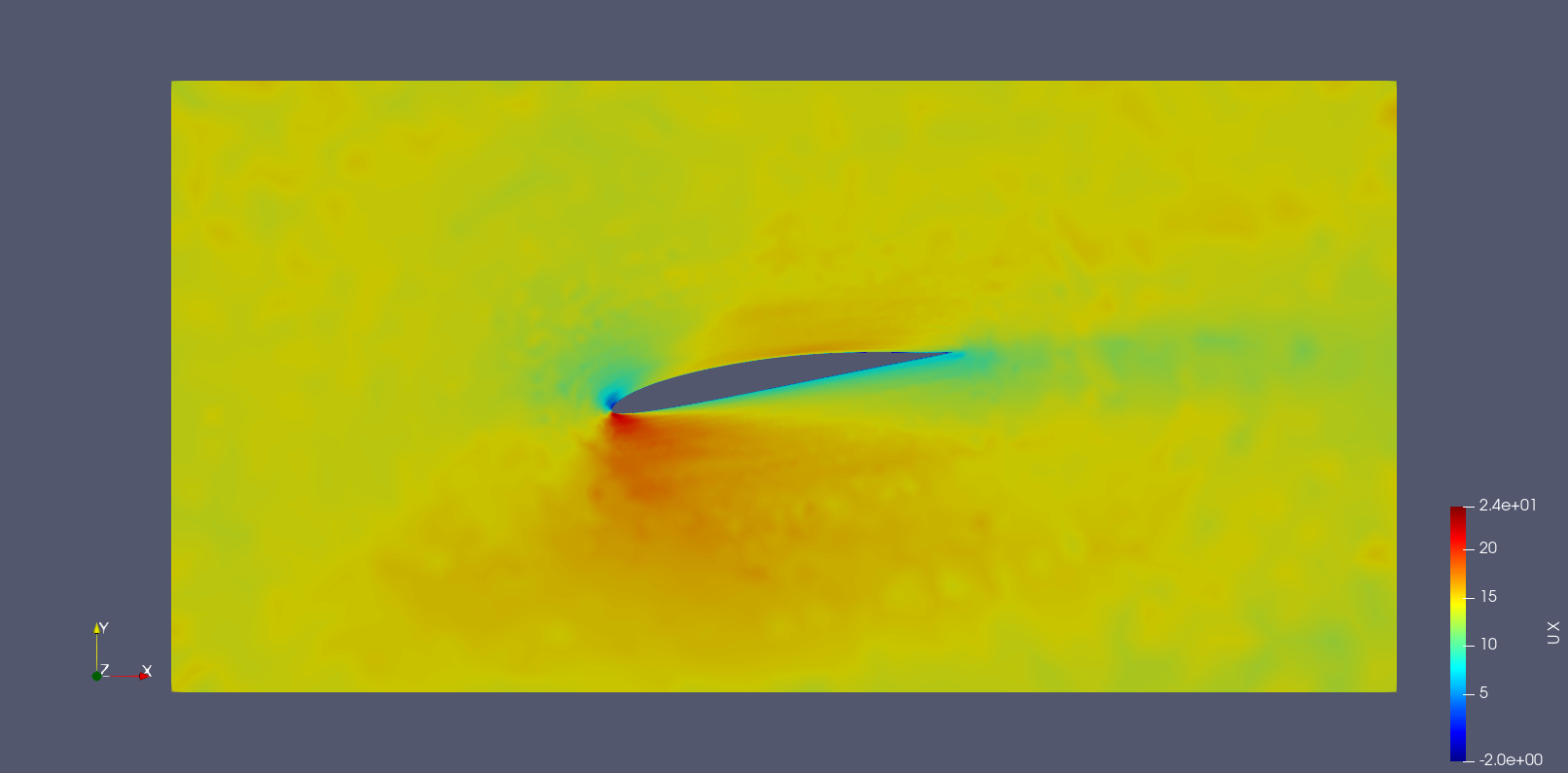}  
      \caption{SIREN kernel}
    \end{subfigure}
    
    \caption{Comparison of the $x$-component of the velocity field for the two different GKN kernels.}
    \label{fig:SIREN}
\end{figure}

\begin{figure}
    \begin{subfigure}{\textwidth}
      \centering
      \includegraphics[width=\linewidth]{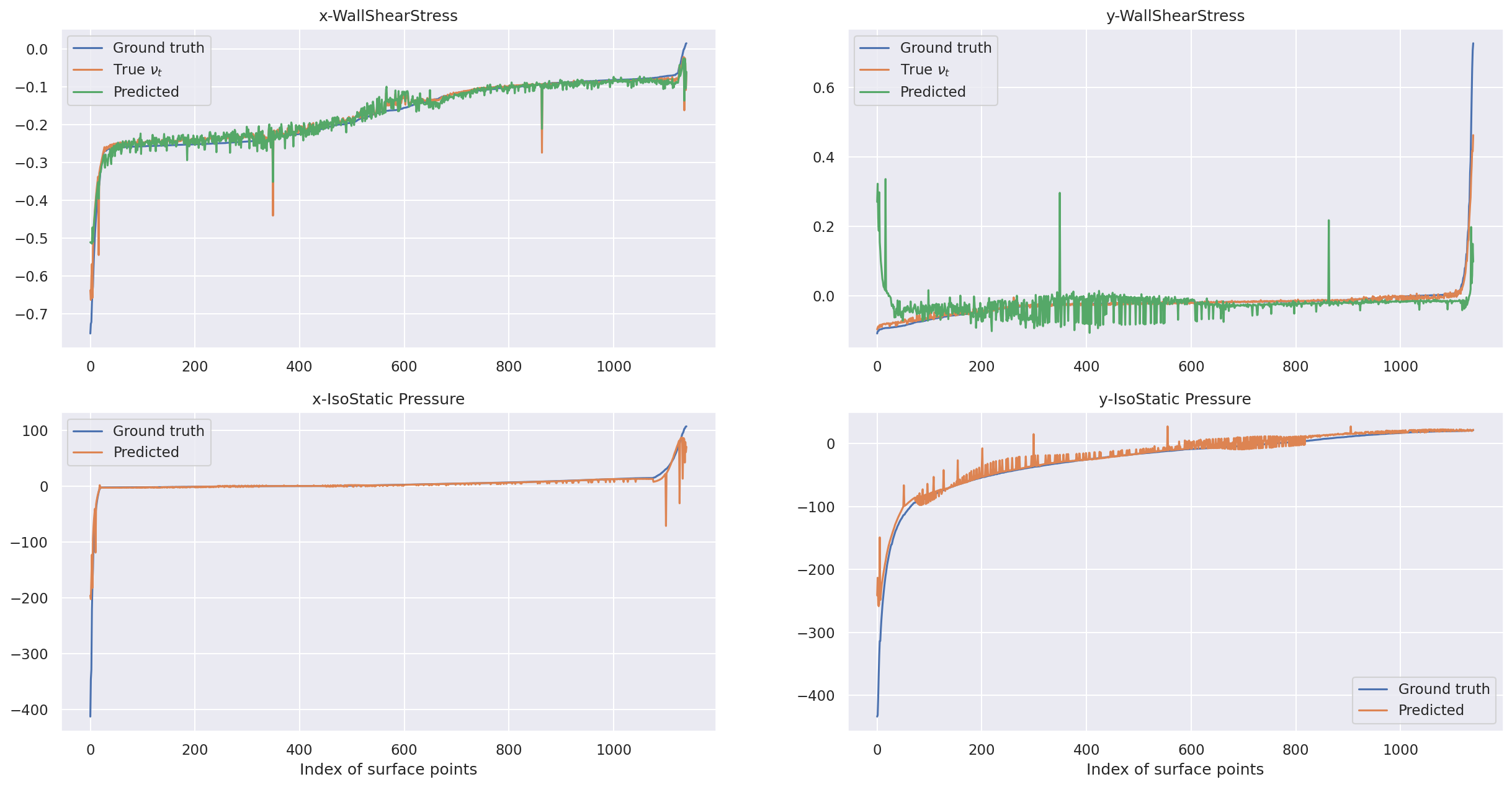}
      \caption{ReLU kernel}
    \end{subfigure}

    \begin{subfigure}{\textwidth}
      \centering
      \includegraphics[width=\linewidth]{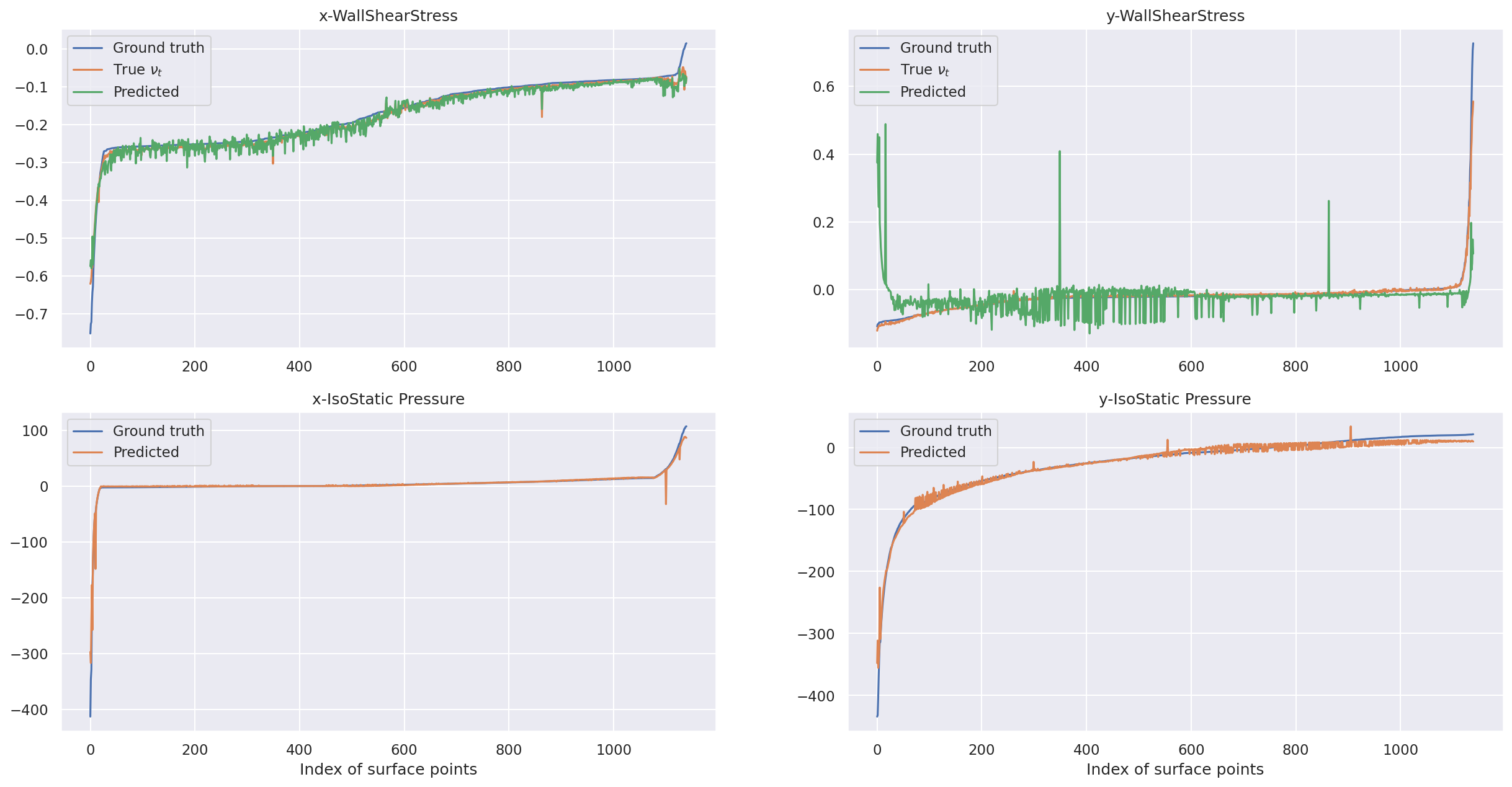}  
      \caption{SIREN kernel}
    \end{subfigure}
    
    \caption{Wall shear stress and isostatic pressure over the surface of the geometry for the two different kernels.}
    \label{fig:SIREN_score}
\end{figure}

Table \ref{SIREN_score} gives the scores of the two kernels, the scores have been generated 10 times and averaged, the standard deviation of those scores is also given, we see that ReLU and SIREN kernels give similar scores. Figure \ref{fig:SIREN} shows the $x$-component of the velocity for the sample with which the models are trained. We see that the SIREN kernel gives oscillatory pattern and has difficulties to learn the low frequencies of the signal whereas the ReLU kernel gives smoother results but less intense signal close to the geometry. Figure \ref{fig:SIREN_score} shows the wall shear stress components and the isostatic pressure components over the surface of the geometry of the same training sample. We again remark that the SIREN kernel is more able to fit sharp signal compare to the ReLU kernel.

Even though the scores of the SIREN kernel are comparable with the ReLU kernel, its ability to fit sharp signal encourage us to find a solution for the overfitting limitation. A recent work \cite{Modulator} propose a way to overcome this limitation and to allow SIREN networks to represent multiples signals and to generalize to new ones. We tried to adapt it to out architecture but we failed to make it work in the validation setting. This means that we were able to train the network over the sampled points and the radius graph but when we use it on a full point clouds of a validation example with a new radius graph, it fails to predict correctly the outputs. The problem is localized in the sampling as it predicts correctly when we do sample in the validation task. This would need more work to make it works but we did not have time to dig further.

We recall that one assumption under the radius graph method was that the interaction between nodes were short-range. This is not the case in generic linear elliptic PDE (and this is actually even not the case in the Poisson equation as we saw in section \ref{PDE}). In order to balance numerical complexity and long-range interactions between nodes, we built a multi scale architecture inspired by the Multipole Graph Kernel Network (MGKN) architecture \cite{MGKN}. The next subsection is dedicated to it. 


\subsubsection{MGKN ReLU}

In this section, we propose a multi-scale architecture to allow long range interactions in the GKN. The idea is to train a GKN for each scale and to make it exchange through a U-Net architecture. The subsampling through the downward pass is done via a uniform sampling and a mean aggregation of the features. This means that each node in the coarser graph represents its closest nodes in the finer graph via a mean aggregation of their features. For the upsampling, the features of a node in the finer graph inherits the same values of the closest node in the coarser graph plus a convolution of the graph at the same scale in the downward pass (except the lower scale that has only the convolution part). More precisely, for $L$ scales, we have at iteration $t$ and for the node $i$: 
\begin{align*}
    \hat{h}_i^{t, l} &= \hat{h}_{k_i}^{t, l+1} + \frac{1}{|\mathcal{N}^l_i|}\sum_{j\in\mathcal{N}^l_i} \kappa_{\Theta_l}(e^{t, l}_{ij})\check{h}^{t, l}_j \qquad l = 1,\,\dots,\, L-1\\
    \hat{h}_i^{t, L} &= \frac{1}{|\mathcal{N}^L_i|}\sum_{j\in\mathcal{N}^L_i} \kappa_{\Theta_L}(e^{t, L}_{ij})\check{h}^{t, L}_j
\end{align*}
where $\check{h}^{t, l}_i$ is the features of the node $i$ at iteration $t$, scale $l$ and in the downward pass, whereas $\hat{h}^{t, l}_i$ is the features of the node $i$ at iteration $t$, scale $l$ but in the upward pass. And where $k_i$ is the closest node to $i$ that is in the coarser graph of scale $l+1$, $\kappa_{\Theta_l}$ is the kernel at scale $l$, $\mathcal{N}_i^l$ is the set of neighbours of node $i$ in the graph at scale $l$ (which includes $i$) and $e_{ij}^l$ is the edge attribute between node $i$ and $j$ in the graph at scale $j$. With those notations, we can write the architecture in the same manner as we did with the GKN, if $h_i\in\mathbb{R}^4$ is our predicted signal at node $i$, $x_i\in\mathbb{R}^6$ our input, $T$ the number of activation and $\sigma$ the activation function:
\begin{align*}
    \begin{cases}
        \check{h}_i^{0, 1} = \phi_\theta(x_i) \\
        \check{h}_i^{t, 1} = \sigma\left(\hat{h}_i^{t-1, 1}\right) + \check{h}_i^{t-1, 1} \qquad t = 1,\,\dots,\, T-1 \\
        h_i = \psi_\gamma\left( \hat{h}_i^{T-1, 1} + \check{h}_i^{T-1, 1} \right)
    \end{cases}
\end{align*}
As in the GKN, we add a residual term in order to help the optimization process.

In our experiments, we chose to set $L$ and $T$ to 5 and to sample $1600-1200-900-600-400$ nodes for the training procedure. The associated radius are $0.1-0.2-0.5-1-10$, where the last one is set enough high to ensure that the radius graph associated is a fully connected graph. Those choices are only motivated by the capacity of our GPU. For the testing procedure, we keep the same radius and the same ratios of subsampling $1-3/4-3/4-2/3-2/3$ and we apply the model over the entire graph. The dimension of the representation is set to 8 and all the 5 kernels are fully connected networks of 8-64-64-64-64 neurons and ReLU activation. The encoder and the decoder are two fully connected networks of 6-64-64-8 and 8-64-64-4 neurons respectively with ReLU activation. The optimization is done with Adam, a one-cycle cosine learning rate of maximum $10^{-3}$, a batch size of 1 and during 1000 epochs.

\begin{figure}
    \begin{subfigure}{\textwidth}
      \centering
      \includegraphics[width=\linewidth]{Baselines/vx_ground_truth.png}
      \caption{Ground truth}
    \end{subfigure}

    \begin{subfigure}{\textwidth}
      \centering
      \includegraphics[width=\linewidth]{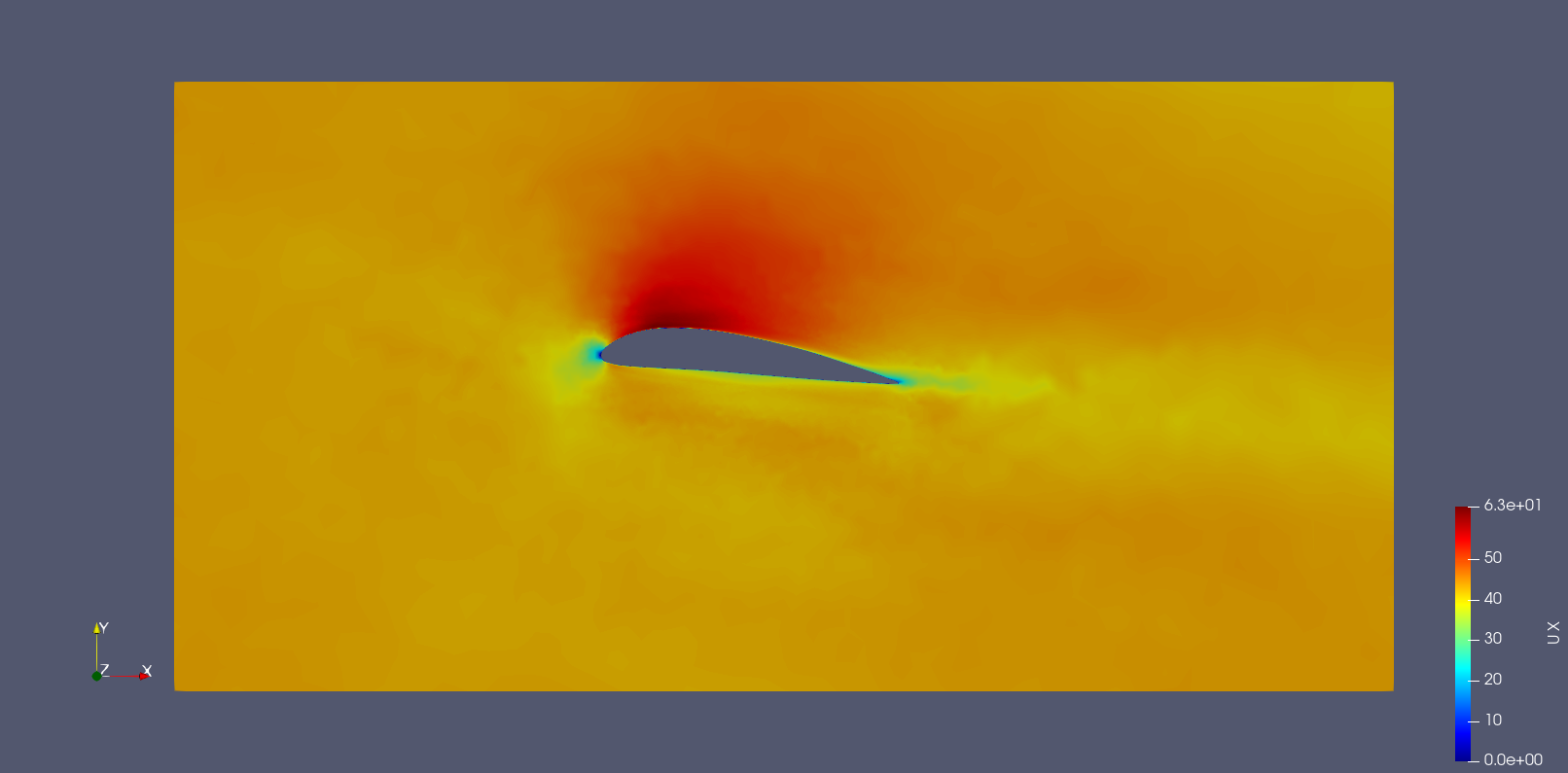}  
      \caption{Predicted}
    \end{subfigure}
    
    \caption{Comparison of the $x$-component of the velocity field for the MGKN with ReLU kernel.}
    \label{fig:MGKN}
\end{figure}

\begin{figure}
    \centering
    \includegraphics[width=\linewidth]{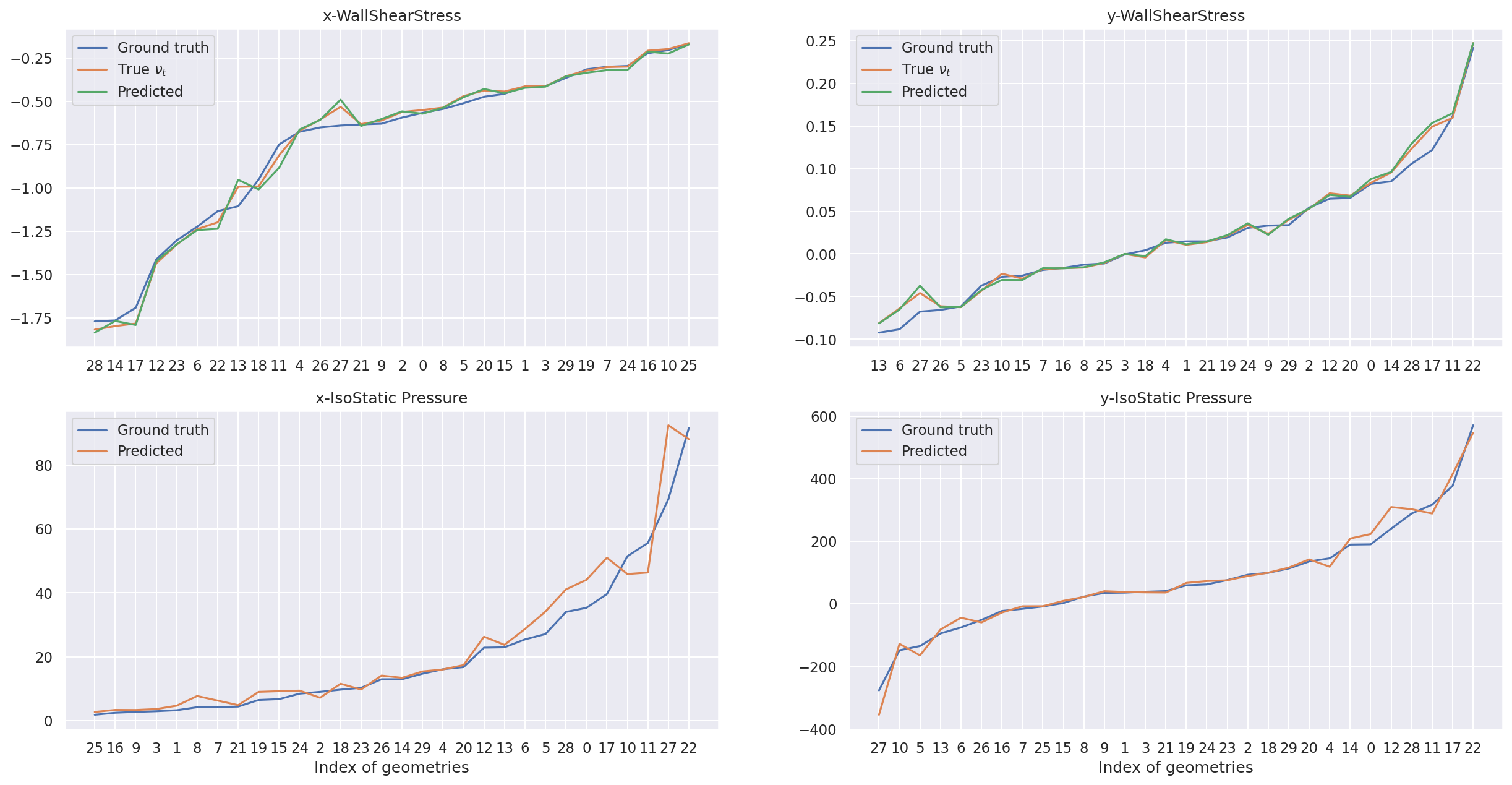}
    \caption{Global coefficients over the test set for the MGKN with ReLU kernel.}
    \label{fig:MGKN_global}
\end{figure}

Table \ref{MGKN_score} gives the scores of the MGKN model. Figure \ref{fig:MGKN} shows the $x$-component of the velocity of an example in the test set and figure \ref{fig:MGKN_global} gives the global coefficients over the entire test set.

\begin{table}
    \centering
    \begin{tabular}{|c|c|c|c|c|c|c|c|}
        \hline
        \multicolumn{2}{|c|}{} & Val & Test & Test noise & Test rot & Test big\footnotemark \\
        \hline
         \multicolumn{2}{|c|}{$\mathcal{L}_\mathcal{V}$} & 0.016 & 0.015 & 0.020 & 0.635 & 0.049 \\
         \multicolumn{2}{|c|}{$\mathcal{L}_\mathcal{S}$} & 0.045 & 0.031 & 0.055 & 0.462 & 0.065 \\
         \cline{1-7}
         \multirow{4}{*}{glob. MAE} & x-WSS & 0.026 & 0.038 & 0.059 & 0.095 & 0.149 \\
         & y-WSS & 0.009 & 0.006 & 0.009 & 0.028 & 0.020 \\
         & x-IsoP & 6.67 & 3.56 & 3.11 & 49.7 & 7.01 \\
         & y-IsoP & 20.7 & 16.7 & 17.6 & 62.2 & 39.5 \\
         \hline
    \end{tabular}
    \caption{Scores of the MGKN model.}
    \label{MGKN_score}
\end{table}

\footnotetext{For this test set, we used the ratios $1-3/8-3/4-2/3-2/3$ for memory footprint.}

Ultimately, we could have replaced the ReLU kernel by a SIREN kernel, this would have led to the same problem as for the GKN, it would not have been possible to train over several samples. Moreover, as we did not manage to make the extension of SIREN work on GKN, we did not try to implement it on a multi-scale architecture.

\subsection{Summary and comparison}

In summary, we built three different models, the GKN with ReLU kernel, the GKN with SIREN kernel and the MGKN with ReLU kernel. Table \ref{total_score} gives a summary of the scores of the different models and baselines over the test set plus the number of parameters and the time of training for each one. 

The GKN largely outperforms the baselines that have only access to the first neighbourhood in the graph (namely the GraphSAGE and the GAT baselines) and achieves similar results compare to the two-scales architecture PointNet with one-third of its number of parameters but is two to three time longer to train. However, even though the GKN is qualitatively better than the GraphSAGE and the GAT baselines when looking at the different fields as images, it creates local artifacts that are not present in the true simulation. In this sense, the PointNet model gives better and smoother qualitative results compare to the GKN ones even though the volumetric MSE is comparable between both models. 

\begin{table}
    \centering
    \begin{tabular}{|c|c|c|c|c|c|c|c|c|}
        \hline
         & $\mathcal{L}_\mathcal{V}$ & $\mathcal{L}_\mathcal{S}$ & x-WSS & y-WSS & x-IsoP & y-IsoP & \#Parameters & Training time \\
        \hline
         GraphSAGE & 0.178 & 0.225 & 0.117 & 0.075 & 10.5 & 57.9 & 17604 & 0:43.47 \\
         GAT & 0.080 & 0.262 & 0.214 & 0.019 & 8.10 & 47.5 & 138532 & 1:14.38  \\
         PointNet & 0.036 & 0.118 & 0.062 & 0.014 & 7.11 & 30.6 & 64892 & 0:47.11 \\
         GKN & 0.039 & 0.048 & 0.040 & 0.008 & 6.52 & 28.2 & 23180 & 2:13.15 \\
         MGKN & \textbf{0.015} & \textbf{0.031} & \textbf{0.038} & \textbf{0.006} & \textbf{3.56} & \textbf{16.7} & 75404 & 36:04.21 \\
         \hline
    \end{tabular}
    \caption{Scores of the different models and baselines over the test set.}
    \label{total_score}
\end{table}

On the other hand, the MGKN model outperforms greatly each of the different models with the same order of magnitude of number of parameters but suffers from a training time thirty to fifty times longer than the other ones. This questions the possible scaling to three dimensional cases even though the training process only use a subsampling of each simulation. However, it produces the best qualitative results for the different fields representation even though it suffers from a patchwork effect inherent of the upsampling procedure in the U-Net. This could be overcome by adding a filter at the end of the U-Net via a new GNN layer.

If we look at the other test sets, the GKN and MGKN models seem to perform better in the generalization setting which incite us to think that they learn more meaningful features and that the few mathematical priors are of good use.

Those observations lead to conclude that the GKN and the MGKN models are good candidates for this task compared to classical out of the box GNN even though the MGKN would surely benefits more engineering to make the training process more efficient and to lower its memory footprint.

\section{Conclusion and Perspectives} \label{conclusion}

In this work, we presented some theoretical results that helped us to understand our task. We adapted architectures in the literature to our problem and leveraged mathematical priors to enhance it and make it more interpretable. Those models largely outperformed basic approaches for such unstructured data leading to good approximation of the global forces acting over the airfoils. We also tried to represent the kernel of our model through an implicit network but failed to make it robust to the downsampling procedure needed in the training step.

We see several perspectives to follow in a next work:
\begin{itemize}[label = \ding{213}]
    \item investigate the failure of the implicit kernel,
    \item leverage the Graph Sampling Based Inductive Learning Method \cite{GraphSAINT} in order to be more robust at the probability density function of the point clouds in inputs,
    \item run new simulations that better respect the incompressibility of the flow and infer a potential vector for the velocity instead of the velocity itself,
    \item do more engineering to enhance the multi-scale structure of the MGKN,
    \item compare those models with other similar approach such as DeepONet \cite{deeponet},
    \item extend to three dimensional cases.
\end{itemize}

\section{Acknowledgement} \label{merci}
Thanks to Extrality for hosting this internship and for their warm inboarding. In particular to the Machine Learning team, Ahmed, Pierre, Louis, Thibaut and Morgane for their reactivity, time and good mood during those six months.

Thanks to the MLIA laboratory and more specially to Patrick Gallinari for the follow-up, the access to the lab and the meetings with PhD students.

Also, thank you Guillaume Charpiat for the discussions, the insights, the friendliness and the Deep Learning in Practice course, they have been of good use during this internship.

And lastly, a warm thanks to Victor, Darius, Malo and Margot for their hospitality. It was so great spending time with you guys!

\bibliographystyle{unsrt}
\bibliography{sample.bib}

\end{document}